\documentclass{article}
\PassOptionsToPackage{numbers,compress}{natbib}
\usepackage[main, final]{neurips_2026}
\usepackage[utf8]{inputenc} % allow utf-8 input
\usepackage[T1]{fontenc}    % use 8-bit T1 fonts
\usepackage{url}            % simple URL typesetting
\usepackage{booktabs}       % professional-quality tables
\usepackage{amsfonts}       % blackboard math symbols
\usepackage{nicefrac}       % compact symbols for 1/2, etc.
\usepackage{microtype}      % microtypography
\usepackage{xcolor}         % colors
\usepackage{amsmath} 
\usepackage{booktabs}
\usepackage{multirow}
\usepackage{pifont}
\usepackage{colortbl}
\usepackage{threeparttable}

\usepackage{xspace}
\usepackage{todonotes}
\usepackage[etex=true,export]{adjustbox}
\usepackage{authblk}    % return to previous author style
\usepackage{makecell}
\usepackage{subcaption}
\usepackage{multirow}
\usepackage{epsfig}
\usepackage{epstopdf}
\usepackage{bbding}
\usepackage{diagbox}
\usepackage{footnote}
\usepackage{color}
\usepackage{xcolor}
\usepackage{wrapfig}
\usepackage{algorithm}
\usepackage{algorithmic}
\usepackage{tabularx}
\usepackage{amsthm}                     
\newtheorem{theorem}{Theorem}          

\definecolor{citecolor}{HTML}{0071bc}
\usepackage[colorlinks=true,linkcolor=red, citecolor=citecolor]{hyperref}
\usepackage[capitalise]{cleveref}

\usepackage{enumitem} 

\title{STDec: Spatio-Temporal Stability Guided Decoding for dLLMs}

\author{%
  \textbf{Yuzhe Chen$^{1}$ \quad Jiale Cao$^{1}$\thanks{Corresponding Author} \quad Xuyang Liu$^{2}$  \quad Jin Xie$^{3}$ \quad Aiping Yang$^{1}$ \quad    Yanwei Pang$^{1}$}
  \\
   $^{1}$ Tianjin University\quad
    $^{2}$ The Hong Kong Polytechnic University\quad
    $^{3}$ Chongqing University\\
    \texttt{\{yzchen,connor,yangaiping,pyw\}@tju.edu.cn} \\ \texttt{xu-yang.liu@connect.polyu.hk \quad xiejin@cqu.edu.cn}
}

\begin{document}
\def\modelname{STDec}

\maketitle

\begin{abstract}
Diffusion Large Language Models (dLLMs) have achieved rapid progress, viewed as a promising alternative to the autoregressive paradigm. However, most dLLM decoders still adopt a global confidence threshold, and do not explicitly model local context from neighboring decoded states or temporal consistency of predicted token IDs across steps. To address this issue, we propose a spatio-temporal stability guided decoding approach, named \modelname{}. We observe strong spatio-temporal stability in dLLM decoding: newly decoded tokens tend to lie near decoded neighbors, and their predicted IDs often remain consistent across several denoising steps. 
Inspired by this stability, our \modelname{} introduces spatial-aware decoding and temporal-aware decoding. The spatial-aware decoding dynamically generates the token-adaptive thresholds by aggregating the decoded states of nearby tokens. The temporal-aware decoding relaxes the decoding thresholds for tokens whose predicted token IDs remain consistent over denoising steps. Our \modelname{} is training-free and remains compatible with cache-based acceleration methods. Across textual reasoning and multimodal understanding benchmarks, \modelname{} achieves improved inference throughput while maintaining comparable generation score. Notably, on MBPP with LLaDA, \modelname{} achieves up to 14.17$\times$ speedup with a comparable score. The homepage is available at  https://yzchen02.github.io/STDec.
\end{abstract}

\section{Introduction}
\label{sec:intro}

Most current LLMs~\cite{gpt4o,llama,qwen3} adopt the autoregressive paradigm, whose left-to-right generation limits inference-time parallelism. Diffusion language models (dLLMs)~\cite{llada,dream} offer a compelling alternative to this left-to-right paradigm. dLLMs can exploit bidirectional context at each refinement step and decode multiple token positions in parallel. Recent dLLMs have shown competitive performance on both textual reasoning~\cite{llada,llada1.5,dream} and multimodal understanding tasks~\cite{mmada,lumina-dimoo,lavida}.

\begin{figure*}[t]
\centering
\includegraphics[width=0.98\linewidth]{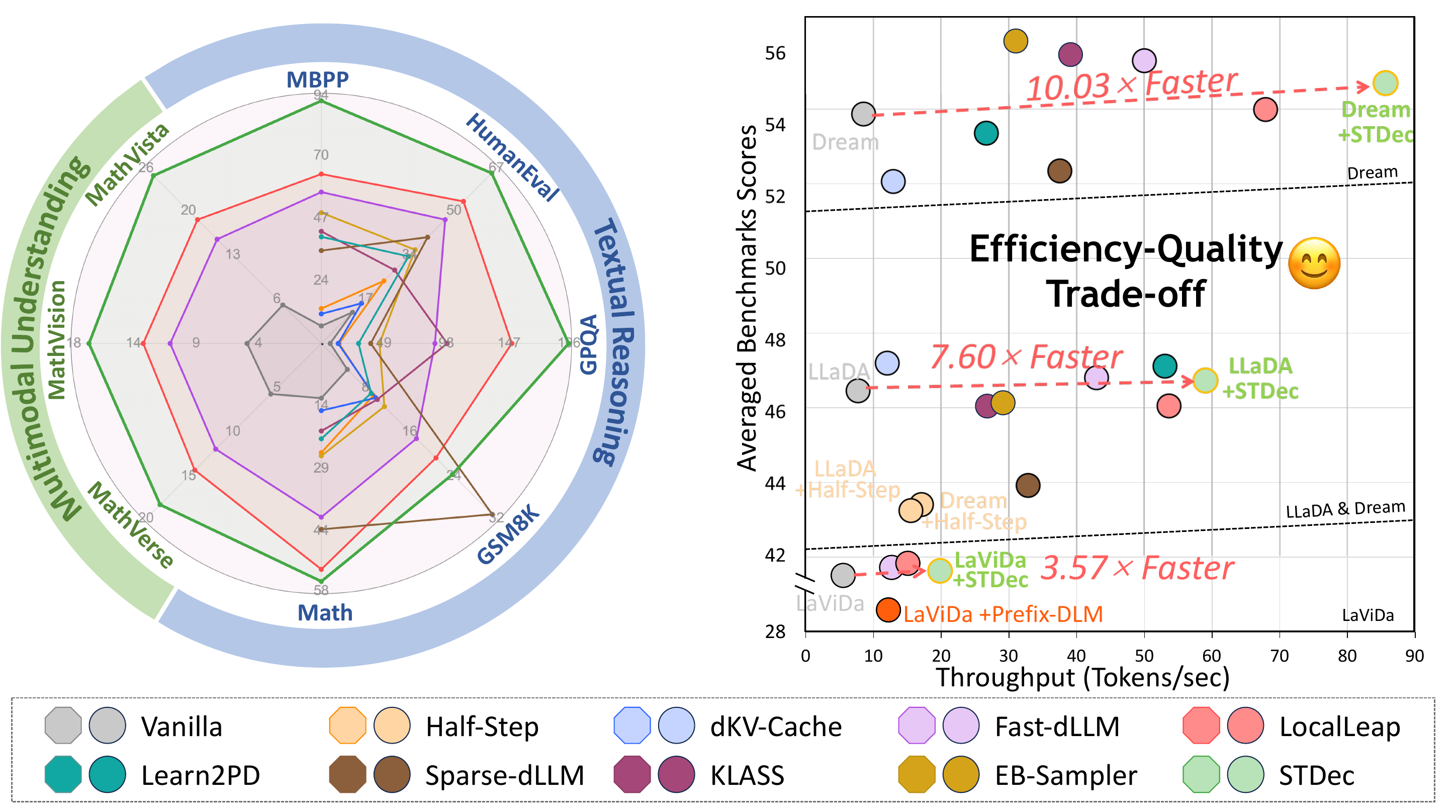}
\caption{\textbf{Efficiency-quality trade-off compared with existing acceleration approaches.} Across eight benchmarks (left) and three models (right), \modelname{} achieves a strong trade-off between throughput and quality, consistently delivering higher generation efficiency while preserving competitive overall performance on both textual reasoning and multimodal understanding tasks.}
\label{fig:intro}
\vspace{-15pt}
\end{figure*}

Although dLLMs are inherently parallel, inference in practical dLLMs~\cite{llada,lavida,lumina-dimoo} remains bottlenecked by conservative decoding policies. Most existing methods finalize a fixed number of high-confidence tokens at each denoising step through a global confidence rule, such as top-$k$ or fixed-threshold decoding. For example, top-1 strategy performs token decoding according to instantaneous confidence, failing to leverage rich spatio-temporal contextual signals. Recent decoding strategies~\cite{localleap,learning-to-parallel,fast-dllm} have considered to incorporate various confidence information of masked tokens for token decoding. However, they still do not jointly exploit the local decoded context and cross-step ID stability as a unified token-level threshold.

We conduct an empirical study to reveal spatio-temporal patterns in dLLM decoding. Spatially, newly decoded tokens tend to emerge near already decoded regions, suggesting that decoded neighbors provide useful cues to perform token decoding. Temporally, the tokens tend to preserve the same predicted IDs at multiple denoising steps before they are finally decoded, indicating that cross-step ID consistency is a common phenomenon before decoded. This motivates us a natural question: \textit{how can we perform token decoding to achieve a better trade-off between efficiency and quality according to the spatio-temporal state?}

In this paper, we propose \modelname{}, a spatio-temporal stability guided decoding method for dLLMs. Our \modelname{} replaces the single global threshold with token-adaptive thresholds according to spatio-temporal stability cues, aiming to improve the efficiency-quality trade-off. Our \modelname{} consists of two complementary components: spatial-aware decoding and temporal-aware decoding. Spatial-aware decoding leverages the local decoded states to generate token-adaptive thresholds. Specifically, initializing from a binary map over decoded and masked positions, we apply a smoothing operation to propagate local decoded-state information, resulting in that masked token closer to more decoded tokens has lower threshold. Temporal-aware decoding further adjusts these thresholds according to historical ID information. When a masked token consistently predicts the same ID across previous denoising steps, we apply a relaxation factor to further lower its threshold. In this way, we can employ spatio-temporal state to jointly guide token-adaptive thresholding for efficient decoding.

We conduct extensive experiments on different textual reasoning and multimodal understanding tasks to validate the effectiveness and efficiency of our \modelname{}. Extensive results indicate that our token-adaptive decoding yields a stronger efficiency-quality trade-off than confidence-only decoding rules. As shown in Fig.~\ref{fig:intro}, our \modelname{} consistently improves decoding throughput while maintaining competitive generation performance on various benchmarks and models. Further, our \modelname{} remains compatible with cache-based acceleration methods. The contributions can be summarized as follows:

\begin{itemize}[leftmargin=12pt, topsep=2pt, itemsep=2pt]
  \item \textbf{Spatio-temporal stability characterization.} We conduct empirical study and show that dLLM decoding exhibits spatio-temporal stability: newly decoded tokens tend to emerge near already decoded regions, and many tokens preserve the same predicted ID across multiple denoising steps before final decoding. These regularities reveal that different tokens can have substantially different decoding states during iterative denoising.

  \item \textbf{Token-adaptive decoding via spatio-temporal cues.} We propose \modelname{}, a training-free spatio-temporal stability guided decoding method that makes token-adaptive decoding based on local decoded states and cross-step ID consistency, enabling faster decoding while maintaining quality.

  \item \textbf{Strong efficiency-quality trade-off.} Extensive experiments on diverse textual reasoning and multimodal understanding benchmarks demonstrate that \modelname{} achieves substantial decoding acceleration while preserving generation quality. Compared with vanilla LLaDA~\cite{llada}, it delivers an average 7.60$\times$ speedup on five textual reasoning benchmarks with no loss in average score. 
\end{itemize}
\section{Related Work}
\subsection{Diffusion Large Language Models}
Discrete diffusion language models (dLLMs) generate text through iterative denoising over masked token sequences, enabling bidirectional context integration and parallel token refinement. Recent dLLMs~\cite{llada,dream,blockdiffusion,tess2} scale this paradigm and achieve competitive results on general language, mathematical reasoning, and code generation benchmarks. Related work explores interpretable latent representations for reasoning~\cite{chang2026struggle,chang2026unlocking}. Beyond model scaling, recent studies~\cite{niescaling,kimtrain,gongscaling,liuthink} have also begun to examine inference-time decoding behavior in dLLMs, including how token commitment order and denoising trajectories affect the quality-efficiency trade-off. In parallel, multimodal dLLMs incorporate visual encoders for vision-language understanding~\cite{lavida,dimple}, while newer architectures further unify discrete diffusion for multimodal understanding and generation~\cite{mmada,lumina-dimoo}. In video understanding, structured temporal reasoning also supports event prediction~\cite{su2026videocoereinforcingvideoevent,Su_2025}. Despite progress in scaling, most dLLM inference pipelines decode tokens by a position-agnostic threshold, leaving local decoded context and cross-step ID stability under-exploited. 

\subsection{Efficient Inference for dLLMs}
Existing acceleration methods for dLLMs mainly follow two directions: reducing per-step cost via caching and increasing per-step progress via decoding policies. Cache-based approaches adapt the KV-cache mechanism to bidirectional diffusion transformers. These methods reuse attention states or intermediate token features across denoising steps in a training-free manner~\cite{dllm-cache,dkv-cache,ma2023deepcache,fast-dllm,zou2024toca}, with further refinements on cache organization, update frequency, feature reuse, and eviction strategies~\cite{flashdlm,sparse-dllm,d2cache,masktokens-prophet}. They effectively reduce redundant computation within each denoising step, but generally do not change the decoding policy that determines which masked tokens are finalized earlier. Other approaches reduce computation through pruning and routing in video models~\cite{wang2026pare,lin2026vcast}. Complementarily, decoding-policy methods accelerate inference by increasing the number of tokens decoded per step or by reducing unnecessary denoising iterations. Some methods exploit local propagation or adaptive parallel decoding to expand decoded regions from confident tokens~\cite{localleap}, while others use phase-wise or slow-fast schedules to allocate different decoding budgets across early and late denoising stages~\cite{slowfast,guo2026phaselora}. Beyond these strategies, other approaches employ convergence-based early stopping~\cite{prophet} or modify denoising and remasking schedules~\cite{remask-scaling,diffusion-forcing,learning-to-parallel,conditional-mask}. Related methods further incorporate inference-time calibration and remasking to mitigate decoding bias~\cite{pc-sampler,dpad,d3tom}. Although these methods improve the efficiency of dLLM inference, many of them still rely on pre-defined schedules, global confidence rules, or single-source trajectory signals, rather than explicitly designing token-level adaptive thresholds from complementary spatial and temporal decoding cues.

Unlike existing dLLM acceleration methods, LocalLeap~\cite{localleap} performs anchor-triggered regional relaxation with a single threshold, while KLASS~\cite{KLASS} uses cross-step distributional stability as a decode-or-defer criterion. In contrast, our STDec constructs token-specific spatial thresholds from the decoded/masked state pattern and further adjusts it using cross-step ID consistency, thereby integrating spatial and temporal cues within a unified token-level threshold. This provides a training-free and model-agnostic decoding rule that increases per-step decoding progress and remains complementary to cache-based acceleration.
\section{Method}

\subsection{Preliminaries}
\label{preliminary}

\paragraph{Diffusion Large Language Models (dLLMs).} dLLMs~\cite{llada,lumina-dimoo} generate sequences by iteratively denoising masked token sequences. dLLMs first encode inputs into a discrete token sequence $X_0 = \{x_0(i)\}_{i=1}^L$, where $L$ is the sequence length and each token belongs to a discrete vocabulary $\mathcal{V}$. A $T$-step Markov masking process gradually corrupts $X_0$ into $X_T$ filled with \texttt{[MASK]} tokens. During inference, a bidirectional Transformer denoiser $p_\theta$ reverses this process. Assuming that  $M_t \subseteq \{1,\dots,L\}$ is the set of masked positions at step $t$,  $\mathcal{D}_t=\{1,\dots,L\}\setminus M_t$ denotes the decoded position, and $P$ are prompt tokens. For each masked position $i\in M_t$, the denoiser predicts a categorical distribution over $\mathcal{V}$:
\begin{equation}
\label{eq:model_dis}
  p_t(i)[v] = p_\theta\!\big(x_t(i)=v \,\big|\, {X_t}, P\big), \quad i \in M_t, v \in \mathcal{V}.
\end{equation}
Then top-1 token can be predicted as $\hat{x}_t(i) = \arg\max_{v\in\mathcal{V}} p_t(i)[v]$ and its confidence is $c_t(i) = p_t(i)\big[\hat{x}_t(i)\big]$.
At each denoising step, the decoder faces a token-level selective decoding decision: whether to decode a masked token now or defer it for further refinement.

\paragraph{Global Threshold based Decoding.} Most existing dLLMs~\cite{llada,lumina-dimoo} adopt a global threshold $\tau_t$ to determine which masked tokens are  decoded at step $t$:
\begin{equation}
  \mathcal{C}_t = \big\{\, i \in M_t \;\big|\; c_t(i) \ge \tau_t \big\},
  \label{eq:topk-threshold}
\end{equation}
where global threshold $\tau_t$ is obtained from top-$k$ strategy~\cite{lumina-dimoo} or fixed threshold~\cite{fast-dllm}.
The masked set is then updated as $M_{t-1}=M_t\setminus \mathcal{C}_t$, and  decoded tokens are fixed as $x(i)=\hat{x}_t(i)$ thereafter. The global-threshold strategy is simple and robust, but it reduces token selection to a confidence-only criterion shared by all masked positions. In iterative denoising, however, global confidence rules do not explicitly use local decoded-state patterns or cross-step prediction consistency, motivating a spatio-temporal reformulation of token-level thresholding.

\begin{figure*}[t]
\centering
\begin{subfigure}[b]{0.48\textwidth}
  \centering
  \includegraphics[width=\textwidth]{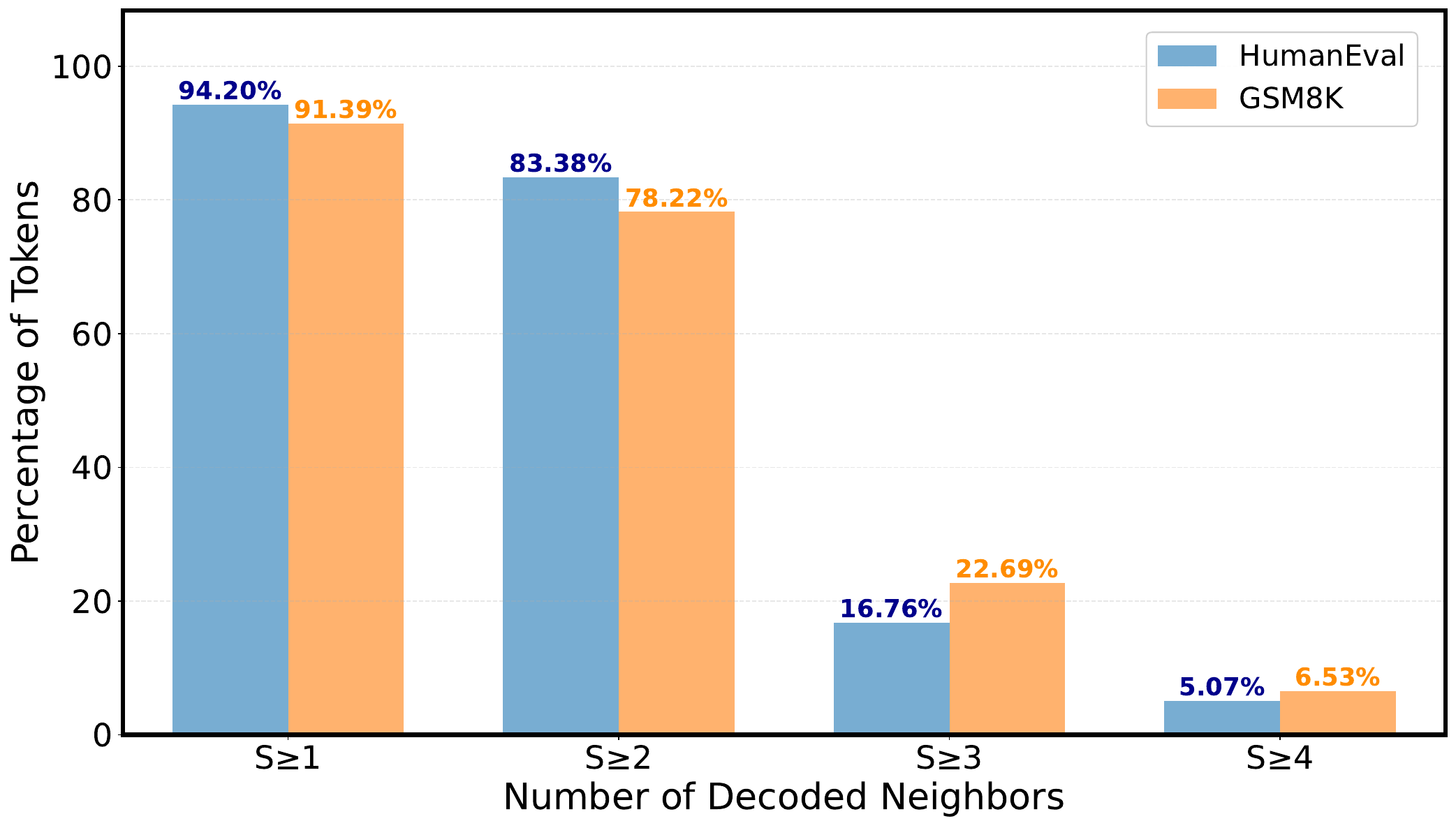}
  \caption{Spatial stability}
  \label{fig:spatial-consistency}
\end{subfigure}
\hfill
\begin{subfigure}[b]{0.48\textwidth}
  \centering
  \includegraphics[width=\textwidth]{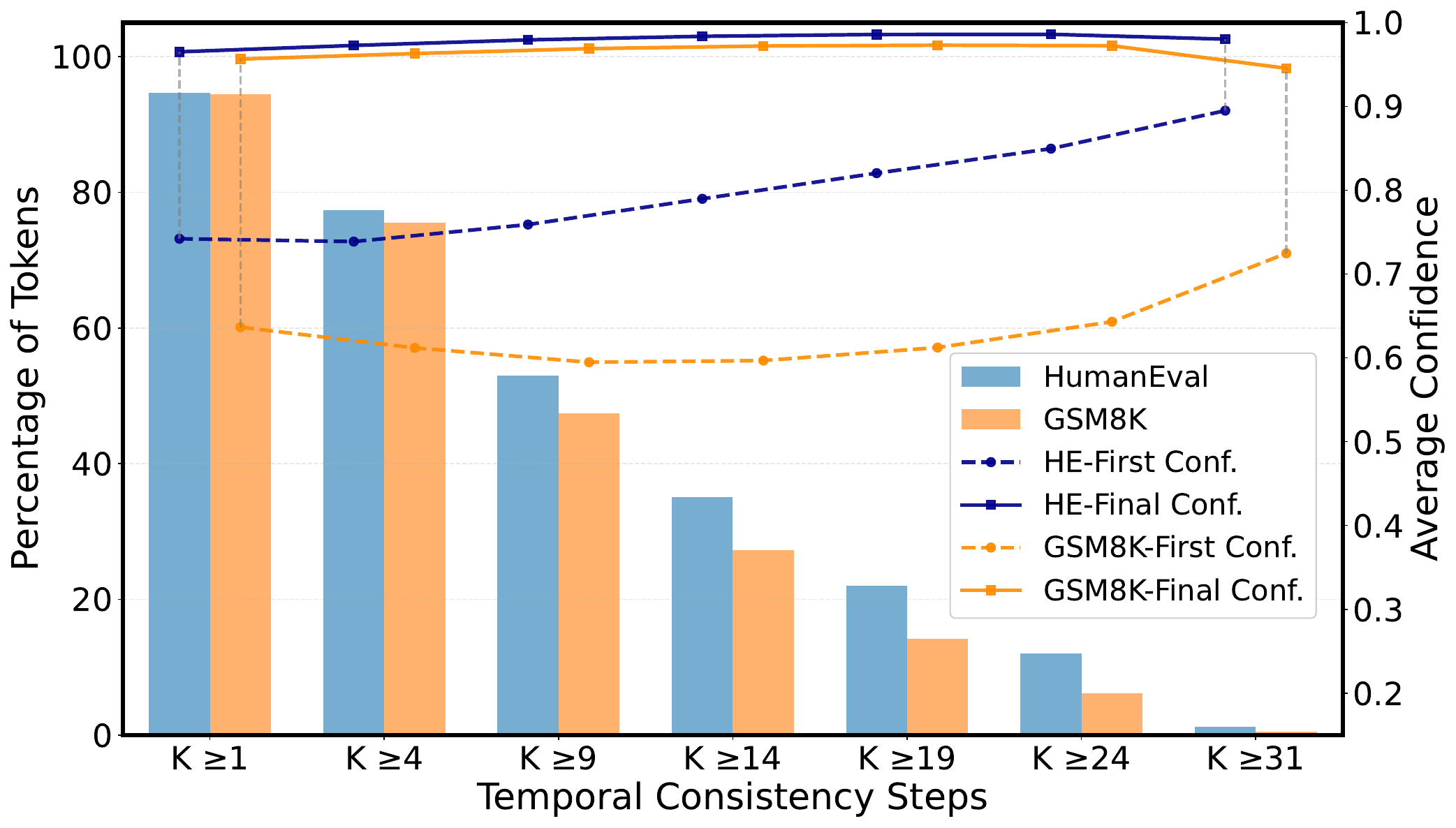}
  \caption{Temporal stability}
  \label{fig:temporal-consistency}
\end{subfigure}
\caption{\textbf{Empirical evidence of spatio-temporal regularities in dLLM decoding.}
(a) \textit{Spatial stability.} For tokens decoded at the current step, we report the percentage whose previously decoded neighbors within a radius of $2$ are at least $S$.
(b) \textit{Temporal stability.} For decoded tokens, we report the percentage whose top-1 predictions have remained unchanged for at least $K$ previous consecutive steps, together with the average confidence at the first ID-stable step and at the final decoded step. Across HumanEval~\cite{humaneval} and GSM8K~\cite{cobbe2021gsm8k}, newly decoded tokens are often located near previously decoded tokens, and many token identities remain consistent before final decoding.}
\label{fig:stability-analysis}
\end{figure*}

\subsection{Empirical Analysis}
\label{sec:motivation}

Here we perform empirical study to reveal there exists spatio-temporal stability in dLLM token decoding, which is used by \modelname{} for token-adaptive thresholding. Specifically, we analyze decoding trajectories on HumanEval~\cite{humaneval} and GSM8K~\cite{cobbe2021gsm8k}, as shown in Fig.~\ref{fig:stability-analysis}.

\paragraph{Spatial Stability.}
We first reveal where newly decoded tokens appear relative to tokens already decoded at previous steps. For each token decoded at the current step, we count the number of previously decoded neighbors within a radius of $2$. As shown in Fig.~\ref{fig:spatial-consistency}, more than $90\%$ of newly decoded tokens have at least one decoded neighbor nearby. Further, among these decoded tokens, a large fraction have multiple decoded neighbors. This indicates that decoded tokens are spatially around already decoded tokens, suggesting that nearby decoded tokens provide a useful local cue for decoding, which can be used to set token-adaptive thresholds.

\paragraph{Temporal Stability.}
We then reveal whether token identities tend to persist before they are decoded. 
For each decoded token, we measure whether its top-1 prediction has remained unchanged for at least $K$ previous consecutive steps. As shown in Fig.~\ref{fig:temporal-consistency}, the fraction is close to $95\%$ for $K{\ge}1$ on both HumanEval and GSM8K. Moreover, the confidence at the first ID-stable step is consistently lower than the confidence at the eventual decoding step. This suggests that cross-step ID consistency captures a complementary temporal signal that is not fully reflected by instantaneous confidence alone, motivating its use as an additional cue for threshold relaxation.

\begin{figure*}[t]
\centering
\includegraphics[width=1.0\linewidth]{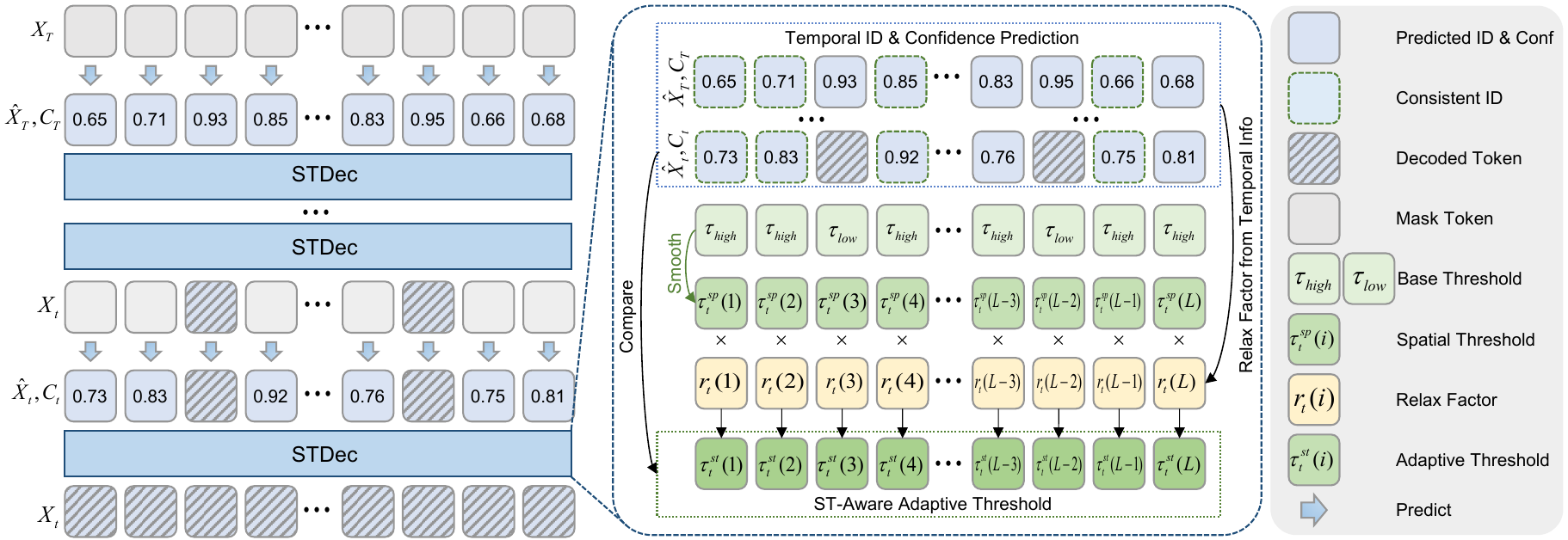}
\caption{\textbf{Overview of our \modelname{}.} The left gives the overall decoding with our \modelname{} in the dLLM, where \modelname{} progressively decodes tokens across denoising steps. The middle gives the details of our \modelname{} that calculates the token-adaptive thresholds by combining spatial and temporal stability.}
\label{fig:method}
\vspace{-13pt}
\end{figure*}

\subsection{Spatio-Temporal Stability Guided Decoding}
\label{sec:stability_decode}

Motivated by the empirical study in Sec.~\ref{sec:motivation}, we introduce a spatio-temporal stability guided decoding strategy that replaces the global threshold with token-adaptive thresholds, named \modelname{}. When designing token-adaptive thresholds, there are two key factors to be considered: \textit{(1) masked tokens around more decoded tokens should receive lower thresholds.} and \textit{(2) cross-step ID consistency can be used to relax threshold.} To achieve this goal, we introduce two complementary modules: spatial-aware decoding and temporal-aware decoding, which adjust the thresholds according to the local decoded state and cross-step ID consistency respectively.

\paragraph{Spatial-aware Decoding.} 
Spatial-aware decoding converts the current decoded/masked state pattern into a spatial threshold for each masked token.
We first choose two seed values $\tau_{\text{low}}<\tau_{\text{high}}$ and construct an auxiliary threshold seed map from the token states:
\begin{equation}
  \tau_t^{\text{init}}(i) =
  \begin{cases}
    \tau_{\text{low}}, & i \notin M_t,\\[2pt]
    \tau_{\text{high}}, & i \in M_t.
  \end{cases}
\end{equation}
The low seed values assigned to decoded positions are not used for decoding decisions on those positions; instead, they are propagated to nearby masked positions through smoothing. 
We apply a Gaussian smoothing operator to the seed map and obtain the spatial threshold for each masked token:
\begin{equation}
  \tau_t^{s}(i) = f_G(\tau_t^{\text{init}}, \sigma)(i), \quad i \in M_t,
\end{equation}
where $\sigma$ controls the smoothing scale and $f_G$ denotes the Gaussian smoothing operation. 
As a result, masked tokens closer to decoded positions receive lower spatial thresholds, whereas tokens in largely masked regions remain closer to $\tau_{\text{high}}$.

\paragraph{Temporal-aware Decoding.} 
The temporal-aware decoding relaxes token thresholds according to historical ID consistency, which serves as an observable proxy for reduced future flip risk during iterative refinement. Let $K_t(i)$ denote the number of consecutive previous steps for which token $i$ keeps the same top-1 predicted ID:
\begin{equation}
K_t(i)=
\begin{cases}
K_{t+1}(i)+1, & \hat{x}_t(i)=\hat{x}_{t+1}(i),\\
0, & \text{otherwise}.
\end{cases}
\end{equation}
We then define a temporal relaxation factor $r_t(i) \in (0,1]$ as
\begin{equation}
  r_t(i) =
  \begin{cases}
    1, & K_t(i) \le 1,\\[2pt]
    \alpha, & K_t(i) > 1,
  \end{cases}
\end{equation}
where $\alpha<1$ is a constant relaxation factor. 
We apply temporal relaxation only when the current top-1 prediction matches those from the two preceding denoising steps, which acts as a conservative guard against rewarding transient or noisy predictions. In practice, we further require the previous-step confidence to exceed $\tau_{\text{low}}$, so that clearly uncertain tokens are excluded from temporal relaxation.

\paragraph{Final Token-Adaptive Decoding.} 
We combine spatial-aware and temporal-aware decoding into a unified token-adaptive decoding strategy: 
\begin{equation}
\tau_t^{st}(i)=\tau_t^s(i)\cdot r_t(i),
\end{equation}
where $\tau_t^s(i)$ denotes the spatial-aware threshold, and $r_t(i)$ is temporal relaxation factor. For the fully masked tokens $X_{T}$, we first predict top-1 IDs $\hat{X_T}=\{\hat{x_T}(i)\}_{i=1}^L$ and corresponding confidence scores $C_T=\{c_T(i)\}_{i=1}^L$.  Based on these confidence scores $C_T$, we employ our proposed \modelname{} to generate token-adaptive thresholds $\tau_T^{st}=\{\tau_T^{st}(i)\}_{i=1}^L$, and then perform token decoding according to these token-adaptive thresholds. The masked tokens are fed into the following step to predict confidence scores and perform decoding again. In this way, we can decode all the masked tokens after T steps.
At step $t$, we can decode the masked tokens as 
\begin{equation}
  \mathcal{C}_t
  = \big\{\, i \in M_t \;\big|\; c_t(i) \ge \tau_t^{\text{st}}(i) \big\},
  \label{eq:commit_set_st}
\end{equation}
which allows different masked tokens to be  decoded under different thresholds at the same step. If no masked token meets its adaptive threshold, we decode the masked token with the highest confidence to ensure progress. We also present the pseudo code of our \modelname{} in Appendix~\ref{sec:decoding}.

\paragraph{Decoding Safety Analysis.}
Here we present theoretical analysis on the decoding safety of token-adaptive decoding.
Following Fast-DLLM~\cite{fast-dllm}, we analyze the decoded positions selected by our \modelname{} under the joint-PMF idealization.
Let $C_t=\{i_1,\ldots,i_n\}$ be the positions decoded at step $t$, and define $\varepsilon_j:=1-c_t(i_j)$ for each $i_j\in C_t$.
The token-adaptive decoding on the selected set $C_t$ yields the same assignment as a greedy sequential decoder operating on  $C_t$ when
\begin{equation}
  \sum_{j=1}^{n}\varepsilon_j+\max_{1\le j\le n}\varepsilon_j<1.
  \label{eq:error-budget}
\end{equation}
This gives a sufficient condition under which the token-adaptive parallel decoding induced by our spatial-aware and temporal-aware modules agrees with ideal greedy sequential decoding on the selected positions.
Appendix~\ref{app:theory} shows the detailed analysis and empirical experiment that about $90\%$ of \modelname{} decoding steps fall within the safe regime.
\section{Experiments}

\subsection{Experiment Setup}
\label{sec:exp_setup}

\noindent\textbf{Baselines.}
For textual reasoning, we compare our \modelname{} with the original decoding strategy in Dream~\cite{dream} and  LLaDA~\cite{llada}, a cache-based method~\cite{dkv-cache,sparse-dllm}, and decoding-policy acceleration methods~\cite{fast-dllm,localleap,learning-to-parallel,KLASS,eb-sampler}. For Learn2PD~\cite{learning-to-parallel}, we only use its policy-only setting. In addition, we include a simple half-step decoding baseline, which directly halves the number of denoising step used. For multimodal understanding, we report a cache-based strategy Prefix-DLM, and some decoding-policy acceleration methods based on LaViDa-Reason~\cite{lavida}.

\noindent\textbf{Implementation Details.} 
For textual reasoning with Dream~\cite{dream} and LLaDA~\cite{llada}, we adopt the semi-autoregressive decoding strategy. For all baselines, we use the hyperparameters recommended in their respective papers. Detailed settings, including generation length, denoising steps, and block size, are provided in Appendix~\ref{sec:evaluation}. For our \modelname{}, we use a single fixed set of stability hyperparameters across all experiments: $\tau_{\text{high}}{=}0.9$, $\tau_{\text{low}}{=}0.3$,  relaxation factor $\alpha{=}0.85$, and Gaussian smoothing with $\sigma{=}1$ and radius 2. These global hyperparameters remain fixed, while the token-level thresholds are dynamically recomputed at each denoising step from the evolving decoding states. All experiments are conducted on NVIDIA RTX 4090D GPUs.

\noindent\textbf{Evaluation Metrics.}
We report task scores (accuracy or Pass@1) and inference throughput, where throughput is measured as the averaged number of output tokens generated per second (TPS). We additionally report Tokens per Forward (TPF), measured as the average number of tokens decoded per model forward pass, and TPS speedup relative to the default decoding strategy in dLLMs.

\subsection{Main Results}
\label{sec:main_results}

\begin{table*}[!t]
\centering
\caption{\textbf{Comparison on textual reasoning benchmarks.} We report TPS, speed, TPF, and score for Dream-7B and LLaDA-8B. Bold values indicate the best TPS, speed, and TPF within each task group, while grayed scores indicate performance drops relative to the corresponding vanilla model.}
\label{tab:textual_reasoning_combined}
\small
\renewcommand{\arraystretch}{0.72}
\setlength{\tabcolsep}{1.35pt}
\begin{tabular*}{\textwidth}{
@{\extracolsep{\fill}}
l
l
c c c c
c c c c
}
\toprule

\multirow{2}{*}{\bfseries Task}
& \multirow{2}{*}{\bfseries Method}
& \multicolumn{4}{c}{\bfseries Dream-7B-Instruct}
& \multicolumn{4}{c}{\bfseries LLaDA-8B-Instruct} \\

\cmidrule(lr){3-6} \cmidrule(lr){7-10}

& &
{\bfseries TPS$\uparrow$} &
{\bfseries Speed$\uparrow$} &
{\bfseries TPF$\uparrow$} &
{\bfseries Score$\uparrow$} &
{\bfseries TPS$\uparrow$} &
{\bfseries Speed$\uparrow$} &
{\bfseries TPF$\uparrow$} &
{\bfseries Score$\uparrow$} \\

\midrule

\multirow{10}{*}{MBPP}
& Vanilla
& 6.57 & 1.00$\times$ & {1.00} & 51.40
& 5.68 & 1.00$\times$ & {1.00} & 37.60 \\

& + Half-Step
& 13.11 & 2.00$\times$ & {2.00} & \textcolor{gray}{35.80}
& 11.51 & 2.03$\times$ & {2.00} & \textcolor{gray}{34.80} \\

& + dKV-Cache~\cite{dkv-cache}
& 11.12 & 1.69$\times$ & {1.00} & \textcolor{gray}{49.60}
& 10.18 & 1.79$\times$ & {1.00} & 39.00 \\

& + Fast-dLLM~\cite{fast-dllm}
& 56.88 & 8.66$\times$ & {8.74} & 54.80
& 62.68 & 11.04$\times$ & {11.42} & \textcolor{gray}{37.40} \\

& + LocalLeap~\cite{localleap}
& 63.65 & 9.69$\times$ & {8.96} & 53.60
& 76.48 & 13.46$\times$ & {11.83} & 37.60 \\

& {+ EB-Sampler~\cite{eb-sampler}}
& {49.14} & {7.48}$\times$ & {7.49} & {55.40}
& {35.58} & {6.26}$\times$ & {6.29} & {38.40} \\

& {+ KLASS~\cite{KLASS}}
& {42.11} & {6.41}$\times$ & {7.48} & {54.80}
& {36.03} & {6.34}$\times$ & {7.02} & \textcolor{gray}{37.40} \\

& + Sparse-dLLM~\cite{sparse-dllm}
& 34.90 & 5.31$\times$ & {1.00} & \textcolor{gray}{48.80}
& 29.83 & 5.25$\times$ & {1.00} & \textcolor{gray}{35.20} \\

& + Learn2PD~\cite{learning-to-parallel}
& 40.07 & 6.10$\times$ & {6.08} & \textcolor{gray}{50.20}
& 76.76 & 13.51$\times$ & {13.35} & 38.00 \\

& \cellcolor{gray!12}\textbf{+ \modelname{}}
& \cellcolor{gray!12}\textbf{91.16}
& \cellcolor{gray!12}\textbf{13.88}$\times$
& \cellcolor{gray!12}{\textbf{14.16}}
& \cellcolor{gray!12}55.60
& \cellcolor{gray!12}\textbf{80.47}
& \cellcolor{gray!12}\textbf{14.17}$\times$
& \cellcolor{gray!12}{\textbf{14.03}}
& \cellcolor{gray!12}38.40 \\

\midrule

% ---------------------- HumanEval --------------------------

\multirow{10}{*}{HumanEval}
& Vanilla
& 11.82 & 1.00$\times$ & {1.00} & 59.15
& 10.82 & 1.00$\times$ & {1.00} & 48.17 \\

& + Half-Step
& 23.69 & 2.00$\times$ & {2.00} & \textcolor{gray}{35.37}
& 21.67 & 2.00$\times$ & {2.00} & \textcolor{gray}{35.37} \\

& + dKV-Cache~\cite{dkv-cache}
& 15.21 & 1.29$\times$ & {1.00} & \textcolor{gray}{56.10}
& 14.23 & 1.32$\times$ & {1.00} & \textcolor{gray}{46.95} \\

& + Fast-dLLM~\cite{fast-dllm}
& 46.87 & 3.97$\times$ & {4.03} & 62.20
& 38.49 & 3.56$\times$ & {3.95} & 48.78 \\

& + LocalLeap~\cite{localleap}
& 53.80 & 4.55$\times$ & {4.46} & \textcolor{gray}{58.54}
& 50.33 & 4.65$\times$ & {\textbf{4.85}} & \textcolor{gray}{46.34} \\

& {+ EB-Sampler~\cite{eb-sampler}}
& {35.48} & {3.00}$\times$ & {3.01} & {63.41}
& {24.47} & {2.26}$\times$ & {2.27} & \textcolor{gray}{43.90} \\

& {+ KLASS~\cite{KLASS}}
& {27.77} & {2.35}$\times$ & {3.07} & {62.20}
& {20.44} & {1.89}$\times$ & {2.31} & {48.17} \\

& + Sparse-dLLM~\cite{sparse-dllm}
& 40.23 & 3.40$\times$ & {1.00} & \textcolor{gray}{58.54}
& 35.93 & 3.32$\times$ & {1.00} & \textcolor{gray}{39.63} \\

& + Learn2PD~\cite{learning-to-parallel}
& 32.99 & 2.79$\times$ & {2.52} & \textcolor{gray}{58.54}
& 51.17 & 4.73$\times$ & {4.77} & \textcolor{gray}{46.95} \\

& \cellcolor{gray!12}\textbf{+ \modelname{}}
& \cellcolor{gray!12}\textbf{64.50}
& \cellcolor{gray!12}\textbf{5.46}$\times$
& \cellcolor{gray!12}{\textbf{5.48}}
& \cellcolor{gray!12}60.37
& \cellcolor{gray!12}\textbf{52.92}
& \cellcolor{gray!12}\textbf{4.89}$\times$
& \cellcolor{gray!12}{\textbf{4.85}}
& \cellcolor{gray!12}48.78 \\

\midrule

% =========================================================
% Mathematics & Science
% =========================================================

% \multicolumn{10}{c}{\bfseries Mathematics \& Science} \\
% \midrule

% ------------------------- GPQA ----------------------------

\multirow{10}{*}{GPQA}
& Vanilla
& 6.95 & 1.00$\times$ & {1.00} & 32.83
& 6.24 & 1.00$\times$ & {1.00} & 28.79 \\

& + Half-Step
& 13.88 & 2.00$\times$ & {2.00} & \textcolor{gray}{32.32}
& 12.66 & 2.03$\times$ & {2.00} & 30.81 \\

& + dKV-Cache~\cite{dkv-cache}
& 13.21 & 1.90$\times$ & {1.00} & 33.33
& 12.33 & 1.98$\times$ & {1.00} & 32.32 \\

& + Fast-dLLM~\cite{fast-dllm}
& 88.95 & 12.80$\times$ & {12.72} & 34.85
& 62.61 & 10.03$\times$ & {12.90} & 28.79 \\

& + LocalLeap~\cite{localleap}
& 149.15 & 21.46$\times$ & {21.83} & 33.33
& 74.60 & 11.96$\times$ & {14.71} & 29.29 \\

& {+ EB-Sampler~\cite{eb-sampler}}
& {45.54} & {6.55}$\times$ & {6.46} & {33.33}
& {57.69} & {9.24}$\times$ & {9.43} & {30.30} \\

& {+ KLASS~\cite{KLASS}}
& {98.54} & {14.18}$\times$ & {14.93} & {32.83}
& {54.34} & {8.71}$\times$ & {9.00} & {29.29} \\

& + Sparse-dLLM~\cite{sparse-dllm}
& 38.66 & 5.56$\times$ & {1.00} & \textcolor{gray}{30.30}
& 34.06 & 5.46$\times$ & {1.00} & 32.83 \\

& + Learn2PD~\cite{learning-to-parallel}
& 29.13 & 4.19$\times$ & {4.11} & 33.33
& 71.03 & 11.38$\times$ & {11.48} & 33.33 \\

& \cellcolor{gray!12}\textbf{+ \modelname{}}
& \cellcolor{gray!12}\textbf{193.62}
& \cellcolor{gray!12}\textbf{27.86}$\times$
& \cellcolor{gray!12}{\textbf{27.94}}
& \cellcolor{gray!12}32.83
& \cellcolor{gray!12}\textbf{92.08}
& \cellcolor{gray!12}\textbf{14.76}$\times$
& \cellcolor{gray!12}{\textbf{14.72}}
& \cellcolor{gray!12}29.29 \\

\midrule

% ------------------------- GSM8K ---------------------------

\multirow{10}{*}{GSM8K}
& Vanilla
& 4.71 & 1.00$\times$ & {1.00} & 83.47
& 4.19 & 1.00$\times$ & {1.00} & 78.01 \\

& + Half-Step
& 9.41 & 2.00$\times$ & {2.00} & \textcolor{gray}{74.22}
& 8.25 & 1.97$\times$ & {2.00} & \textcolor{gray}{75.82} \\

& + dKV-Cache~\cite{dkv-cache}
& 9.83 & 2.09$\times$ & {1.00} & \textcolor{gray}{79.08}
& 8.82 & 2.11$\times$ & {1.00} & \textcolor{gray}{77.63} \\

& + Fast-dLLM~\cite{fast-dllm}
& 17.20 & 3.65$\times$ & {3.67} & \textcolor{gray}{82.94}
& 12.16 & 2.90$\times$ & {3.16} & 78.77 \\

& + LocalLeap~\cite{localleap}
& 20.71 & 4.40$\times$ & {4.45} & \textcolor{gray}{82.49}
& 17.01 & 4.06$\times$ & {\textbf{4.02}} & \textcolor{gray}{77.98} \\

& {+ EB-Sampler~\cite{eb-sampler}}
& {11.40} & {2.42}$\times$ & {2.42} & {83.70}
& {7.64} & {1.82}$\times$ & {1.84} & {78.77} \\

& {+ KLASS~\cite{KLASS}}
& {10.12} & {2.15}$\times$ & {2.28} & {83.55}
& {8.97} & {2.14}$\times$ & {2.26} & \textcolor{gray}{76.80} \\

& + Sparse-dLLM~\cite{sparse-dllm}
& \textbf{30.90} & \textbf{6.56}$\times$ & {1.00} & \textcolor{gray}{81.88}
& \textbf{26.11} & \textbf{6.23}$\times$ & {1.00} & \textcolor{gray}{77.71} \\

& + Learn2PD~\cite{learning-to-parallel}
& 9.01 & 1.91$\times$ & {1.83} & \textcolor{gray}{82.56}
& 16.49 & 3.94$\times$ & {3.97} & 78.09 \\

& \cellcolor{gray!12}\textbf{+ \modelname{}}
& \cellcolor{gray!12}23.70
& \cellcolor{gray!12}5.03$\times$
& \cellcolor{gray!12}{\textbf{5.01}}
& \cellcolor{gray!12}\textcolor{gray}{82.34}
& \cellcolor{gray!12}16.94
& \cellcolor{gray!12}4.04$\times$
& \cellcolor{gray!12}{4.00}
& \cellcolor{gray!12}78.01 \\

\midrule

% ------------------------- MATH ----------------------------

\multirow{10}{*}{MATH}
& Vanilla
& 12.67 & 1.00$\times$ & {1.00} & 44.64
& 11.90 & 1.00$\times$ & {1.00} & 40.38 \\

& + Half-Step
& 25.31 & 2.00$\times$ & {2.00} & \textcolor{gray}{39.40}
& 23.64 & 1.99$\times$ & {2.00} & \textcolor{gray}{39.44} \\

& + dKV-Cache~\cite{dkv-cache}
& 15.56 & 1.23$\times$ & {1.00} & \textcolor{gray}{44.04}
& 14.73 & 1.24$\times$ & {1.00} & 40.90 \\

& + Fast-dLLM~\cite{fast-dllm}
& 40.26 & 3.18$\times$ & {3.22} & \textcolor{gray}{44.16}
& 39.00 & 3.28$\times$ & {3.54} & 41.00 \\

& + LocalLeap~\cite{localleap}
& 52.35 & 4.13$\times$ & {4.28} & \textcolor{gray}{44.26}
& 49.93 & 4.20$\times$ & {\textbf{4.65}} & \textcolor{gray}{39.66} \\

& {+ EB-Sampler~\cite{eb-sampler}}
& {26.05} & {2.06}$\times$ & {2.06} & {48.22}
& {22.38} & {1.88}$\times$ & {1.89} & {41.14} \\

& {+ KLASS~\cite{KLASS}}
& {20.28} & {1.60}$\times$ & {2.13} & {45.90}
& {23.75} & {2.00}$\times$ & {2.50} & \textcolor{gray}{38.94} \\

& + Sparse-dLLM~\cite{sparse-dllm}
& 43.09 & 3.40$\times$ & {1.00} & \textcolor{gray}{44.08}
& 38.49 & 3.23$\times$ & {1.00} & \textcolor{gray}{34.44} \\

& + Learn2PD~\cite{learning-to-parallel}
& 22.09 & 1.74$\times$ & {1.47} & \textcolor{gray}{44.26}
& 49.88 & 4.19$\times$ & {4.23} & \textcolor{gray}{40.04} \\

& \cellcolor{gray!12}\textbf{+ \modelname{}}
& \cellcolor{gray!12}\textbf{55.15}
& \cellcolor{gray!12}\textbf{4.35}$\times$
& \cellcolor{gray!12}{\textbf{4.45}}
& \cellcolor{gray!12}44.64
& \cellcolor{gray!12}\textbf{53.05}
& \cellcolor{gray!12}\textbf{4.46}$\times$
& \cellcolor{gray!12}{4.58}
& \cellcolor{gray!12}\textcolor{gray}{39.86} \\

\midrule
\midrule

% =========================================================
% Average
% =========================================================

\multirow{10}{*}{\textit{Avg.}}
& Vanilla
& 8.54 & 1.00$\times$ & {1.00} & 54.30
& 7.77 & 1.00$\times$ & {1.00} & 46.59 \\

& + Half-Step
& 17.08 & 2.00$\times$ & {2.00} & \textcolor{gray}{43.42}
& 15.55 & 2.00$\times$ & {2.00} & \textcolor{gray}{43.25} \\

& + dKV-Cache~\cite{dkv-cache}
& 12.99 & 1.52$\times$ & {1.00} & \textcolor{gray}{52.43}
& 12.06 & 1.55$\times$ & {1.00} & 47.36 \\

& + Fast-dLLM~\cite{fast-dllm}
& 50.03 & 5.86$\times$ & {6.47} & 55.79
& 42.99 & 5.53$\times$ & {6.99} & 46.95 \\

& + LocalLeap~\cite{localleap}
& 67.93 & 7.95$\times$ & {8.80} & \textcolor{gray}{54.44}
& 53.67 & 6.91$\times$ & {8.01} & \textcolor{gray}{46.17} \\

& {+ EB-Sampler~\cite{eb-sampler}}
& {33.52} & {3.93}$\times$ & {4.29} & {56.81}
& {29.55} & {3.80}$\times$ & {4.34} & \textcolor{gray}{46.50} \\

& {+ KLASS~\cite{KLASS}}
& {39.76} & {4.66}$\times$ & {5.98} & {55.86}
& {28.71} & {3.69}$\times$ & {4.62} & \textcolor{gray}{46.12} \\

& + Sparse-dLLM~\cite{sparse-dllm}
& 37.56 & 4.40$\times$ & {1.00} & \textcolor{gray}{52.72}
& 32.88 & 4.23$\times$ & {1.00} & \textcolor{gray}{43.96} \\

& + Learn2PD~\cite{learning-to-parallel}
& 26.66 & 3.12$\times$ & {3.20} & \textcolor{gray}{53.78}
& 53.07 & 6.83$\times$ & {7.56} & 47.28 \\

& \cellcolor{gray!12}\textbf{+ \modelname{}}
& \cellcolor{gray!12}\textbf{85.63}
& \cellcolor{gray!12}\textbf{10.03}$\times$
& \cellcolor{gray!12}{\textbf{11.41}}
& \cellcolor{gray!12}55.16
& \cellcolor{gray!12}\textbf{59.09}
& \cellcolor{gray!12}\textbf{7.60}$\times$
& \cellcolor{gray!12}{\textbf{8.43}}
& \cellcolor{gray!12}46.87\\

\bottomrule
\end{tabular*}
\vspace{-7mm}
\end{table*}

\noindent\textbf{Performance on Dream.} 
Table~\ref{tab:textual_reasoning_combined} gives the results on five different benchmarks based on Dream~\cite{dream}. On these benchmarks, our \modelname{} achieves higher TPS while maintaining comparable scores with original Dream. Specifically, our \modelname{} improves TPS from 6.57 to 91.16 on MBPP ($13.88\times$) and from 11.82 to 64.50 on HumanEval ($5.46\times$). On mathematics and science benchmarks, our \modelname{} yields substantial gains in decoding speed, such as $5.03\times$ on GSM8K, $4.35\times$ on MATH, and $27.86\times$ on GPQA, while almost maintaining the scores. 
When simply halving the number of denoising steps, it presents $2\times$ speedup but leads to a significant drop in scores on most benchmarks, demonstrating that simple step reduction cannot simultaneously accelerate and maintain its quality.
Compared to other methods~\cite{dkv-cache,fast-dllm,localleap,sparse-dllm,learning-to-parallel}, our \modelname{} offers a better efficiency–quality trade-off.  

\noindent\textbf{Performance on LLaDA.} Table~\ref{tab:textual_reasoning_combined} further presents the results on textual reasoning benchmarks with LLaDA~\cite{llada}. Similarly, our \modelname{} gives the best trade-off between TPS and quality.  For code generation, our \modelname{} increases TPS from 5.68 to 80.47 on MBPP and from 10.82 to 52.92 on HumanEval, without loss of quality. For mathematics and science, our \modelname{} obtains $4.04\times$ speedup on GSM8K, $4.46\times$ on MATH, and $14.76\times$ on GPQA, while almost preserving generation scores. Compared to Fast-dLLM~\cite{fast-dllm} and Learn2PD~\cite{learning-to-parallel}, our \modelname{} achieves higher average throughput. In terms of TPF, STDec achieves an average TPF of 11.41 on Dream~\cite{dream} and 8.43 on LLaDA~\cite{llada}. These values are substantially higher than those of KLASS~\cite{KLASS} (5.98 and 4.62) and EB-Sampler~\cite{eb-sampler} (4.29 and 4.34), while STDec maintains comparable average task performance.

\textbf{Open-ended generation.}
We further evaluate STDec on 183 English and Chinese creative-writing prompts from WritingBench~\cite{WritingBench}, using the default hyperparameters without task-specific tuning. STDec improves TPS from 4.36 to 7.99 (1.83$\times$) and TPF from 1.00 to 1.77, while maintaining comparable English and Chinese scores.

\begin{table}[t]
\centering
\renewcommand{\arraystretch}{1.0}
\setlength{\tabcolsep}{2.8pt}

\caption{\textbf{Comparison on multimodal understanding with LaViDa-Reason.} We report TPS, speedup, and score across benchmarks. Bold values indicate the best efficiency, and grayed scores denote performance drop.}
\label{tab:lavida_table}

\small
\begin{tabular*}{\columnwidth}{@{\extracolsep{\fill}} l|ccc|ccc|ccc|ccc}
\toprule
& \multicolumn{3}{c|}{MathVerse}
& \multicolumn{3}{c|}{MathVision}
& \multicolumn{3}{c|}{MathVista}
& \multicolumn{3}{c}{\textit{Avg.}} \\
\cmidrule(r){2-4}
\cmidrule(r){5-7}
\cmidrule(r){8-10}
\cmidrule(l){11-13}

\textbf{Method}
& TPS & Speed & Score
& TPS & Speed & Score
& TPS & Speed & Score
& TPS & Speed & Score \\
\midrule

\shortstack[l]{LaViDa~\cite{lavida}}
& 5.71 & 1.00$\times$ & 28.30
& 5.34 & 1.00$\times$ & 19.74
& 5.66 & 1.00$\times$ & 47.20
& 5.57 & 1.00$\times$ & 31.75 \\

+ Prefix-DLM~\cite{lavida}
& 12.22 & 2.14$\times$ & \textcolor{gray}{27.03}
& 12.09 & 2.26$\times$ & 20.39
& 12.38 & 2.19$\times$ & \textcolor{gray}{40.90}
& 12.23 & 2.20$\times$ & \textcolor{gray}{29.44} \\

\shortstack[l]{+ Fast-dLLM~\cite{fast-dllm}}
& 11.94 & 2.09$\times$ & 28.68
& 10.87 & 2.04$\times$ & 20.72
& 15.31 & 2.70$\times$ & 47.50
& 12.71 & 2.28$\times$ & 32.30 \\

+ LocalLeap~\cite{localleap}
& 14.31 & 2.51$\times$ & 28.30
& 12.81 & 2.40$\times$ & 21.71
& 18.19 & 3.21$\times$ & 47.70
& 15.10 & 2.71$\times$ & 32.57 \\

\rowcolor{gray!10}
\textbf{+ \modelname{}}
& \textbf{18.22} & \textbf{3.19}$\times$ & 28.30
& \textbf{16.71} & \textbf{3.13}$\times$ & 21.71
& \textbf{24.66} & \textbf{4.36}$\times$ & \textcolor{gray}{46.20}
& \textbf{19.86} & \textbf{3.57}$\times$ & 32.07 \\

\bottomrule
\end{tabular*}

\vspace{-2mm}
\end{table}

\noindent\textbf{Performance on LaViDa.} Table~\ref{tab:lavida_table} reports the results of different acceleration strategies with LaViDa-Reason~\cite{lavida} on multimodal understanding tasks. Our \modelname{} also provides a good efficiency-quality trade-off on multimodal understanding. Specifically, our \modelname{} improves TPS from 5.57 to 19.86  ($3.57\times$) averaged over three benchmarks. Compared to the built-in cache-based and other sampling-based approaches, our \modelname{} consistently achieves higher TPS on all benchmarks maintaining comparable performance.

\begin{figure}[t]
\centering
\footnotesize
\begin{subfigure}[b]{0.325\textwidth}
  \centering
  \includegraphics[width=\textwidth]{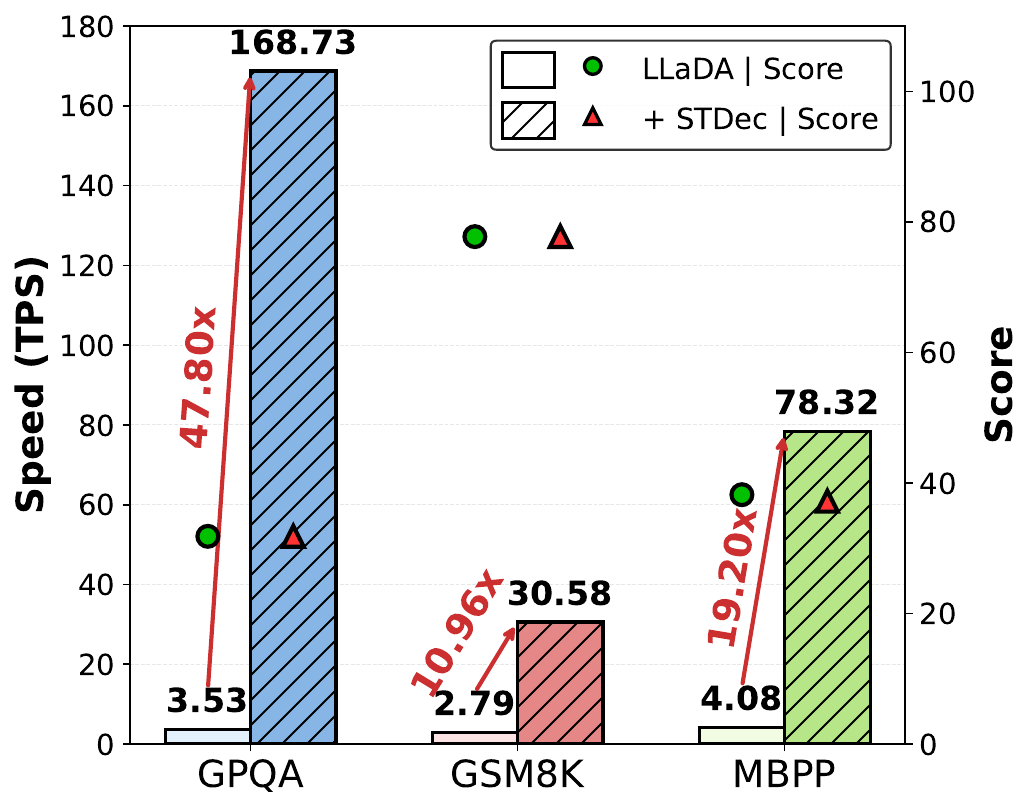}
  \caption{}
  \label{fig:speed-bars-third}
\end{subfigure}
\begin{subfigure}[b]{0.32\textwidth}
  \centering
  \includegraphics[width=\textwidth]{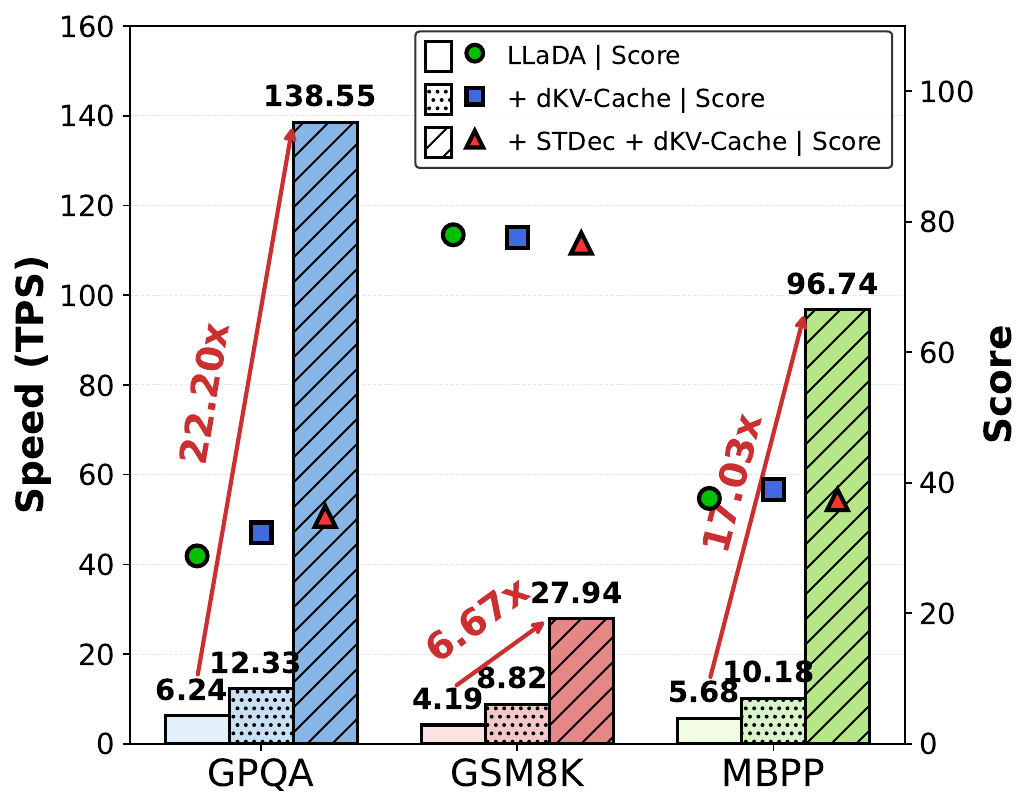}
  \caption{}
  \label{fig:speed-bars-llada}
\end{subfigure}\hfill
\begin{subfigure}[b]{0.32\textwidth}
  \centering
  \includegraphics[width=\textwidth]{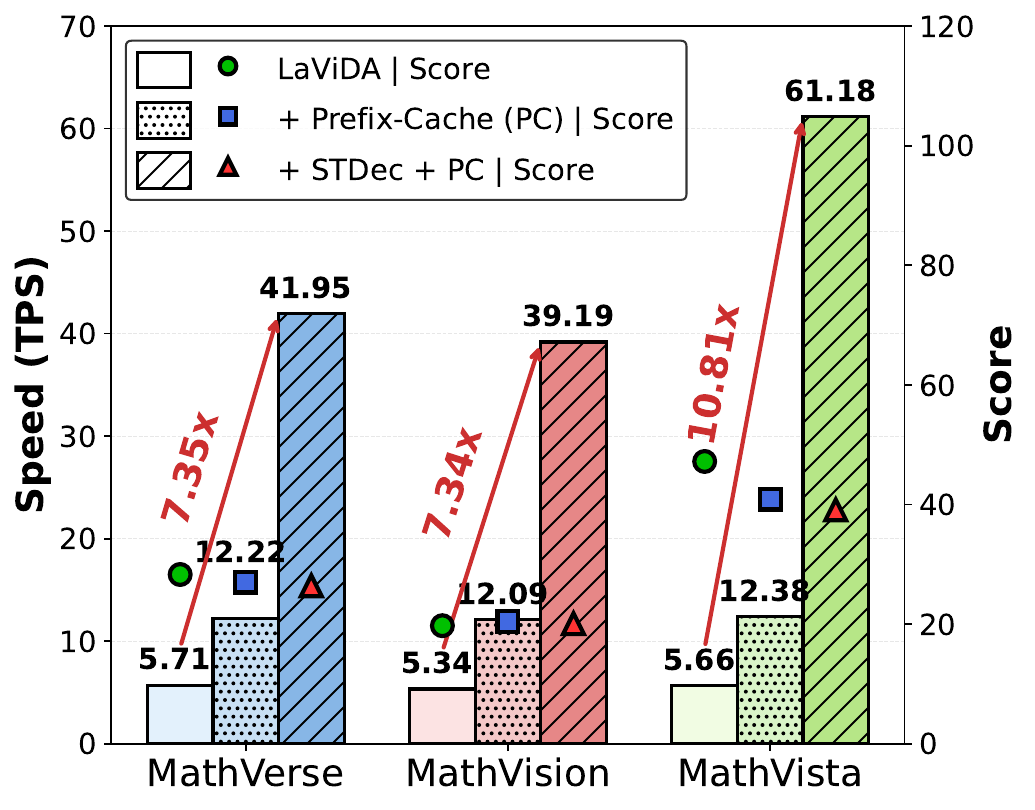}
  \caption{}
  \label{fig:speed-bars-lavida}
\end{subfigure}\hfill
\vspace{-1mm}
\caption{\textbf{Further scalability of our \modelname{}.} \textit{(a) Long-context textual reasoning.} Our \modelname{} can be adopted for long-context generation  (1024 tokens). \textit{(b-c) Combination with cache-based approaches.} Our method can be integrated with cache-based approaches dKV-Cache \cite{dkv-cache} and Prefix-DLM \cite{lavida}, which can accelerate the inference speed almost without loss of generation quality.}
\vspace{-20pt}
\label{fig:speed-bars}
\end{figure}

\noindent\textbf{Further Scalability.}
To further validate the efficiency of our \modelname{}, we evaluate its scalability along two aspects: long-context textual reasoning and compatibility with cache-based acceleration.

\textit{(i) Long-context textual reasoning.}
Fig.~\ref{fig:speed-bars-third} reports 1024-token generation results with LLaDA~\cite{llada}.
Our \modelname{} consistently improves throughput by $10.96\times$-$47.80\times$ with comparable scores; for example, it achieves a $47.80\times$ speedup on GPQA with the same score.
These results suggest that \modelname{} scales favorably to long-context decoding, substantially improving throughput while keeping performance broadly comparable.

\textit{(ii) Stacking \modelname{} with cache-based acceleration.}
Fig.~\ref{fig:speed-bars-llada} and Fig.~\ref{fig:speed-bars-lavida} show that \modelname{} is complementary to caching-based approaches. This complementarity arises because caching reduces the cost of each forward pass, whereas STDec reduces the required number of forward passes. On textual reasoning with LLaDA~\cite{llada}, stacking \modelname{} on top of dKV-Cache~\cite{dkv-cache} consistently brings additional gains over the cache baseline: $11.2\times$ on GPQA, $3.2\times$ on GSM8K, and $9.5\times$ on MBPP in Fig.~\ref{fig:speed-bars-llada}.
On multimodal understanding with LaViDa~\cite{lavida}, which natively supports Prefix-DLM caching, adding \modelname{} on top of Prefix-DLM yields similar speed improvements while maintaining comparable performance in Fig.~\ref{fig:speed-bars-lavida}.

We further extend our \modelname{} to image generation, additional scalability results are provided in Appendix~\ref{sec:extension}.
These additional experiments demonstrate that \modelname{} is robust across sequence lengths and modalities, and can be stacked with cache-based methods.

\subsection{Ablation Studies}
\label{sec:ablation}

We conduct extensive ablation studies on HumanEval~\cite{humaneval} with LLaDA-8B-Instruct~\cite{llada}. In addition, we present some additional ablation experiments in Appendix~\ref{sec:additional-ablation}.

\begin{wrapfigure}{r}{0.32\textwidth}
    \centering
    \includegraphics[width=\linewidth]{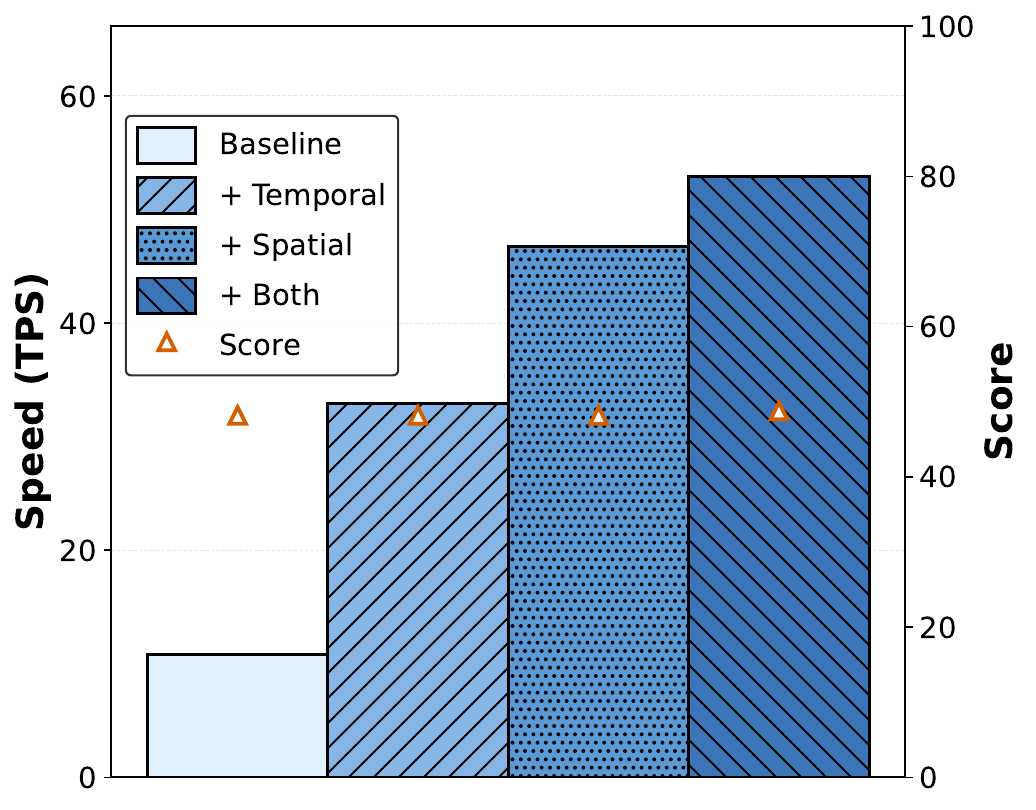}
    \caption{\textbf{Impact of integrating our STDec modules.} We integrate our spatial-aware decoding and temporal-aware decoding into the baseline LLaDA~\cite{llada}.}
    \label{fig:module-ablation}
\end{wrapfigure}

\noindent\textbf{Impact of Spatial-aware and Temporal-aware Decoding.} Fig.~\ref{fig:module-ablation} presents the results of integrating our proposed modules into the baseline LLaDA~\cite{llada}. When only integrating the temporal-aware decoding into the baseline, it improves TPS from 10.82 to 32.89 without sacrificing score. When integrating these two modules, it provides the fastest speed of 52.92 TPS. Namely, our spatial-aware decoding and temporal-aware decoding are complementary, together enabling faster speed.

\begin{figure}[t]
\centering
\footnotesize
\begin{subfigure}[b]{0.244\textwidth}
  \centering
  \includegraphics[width=\textwidth]{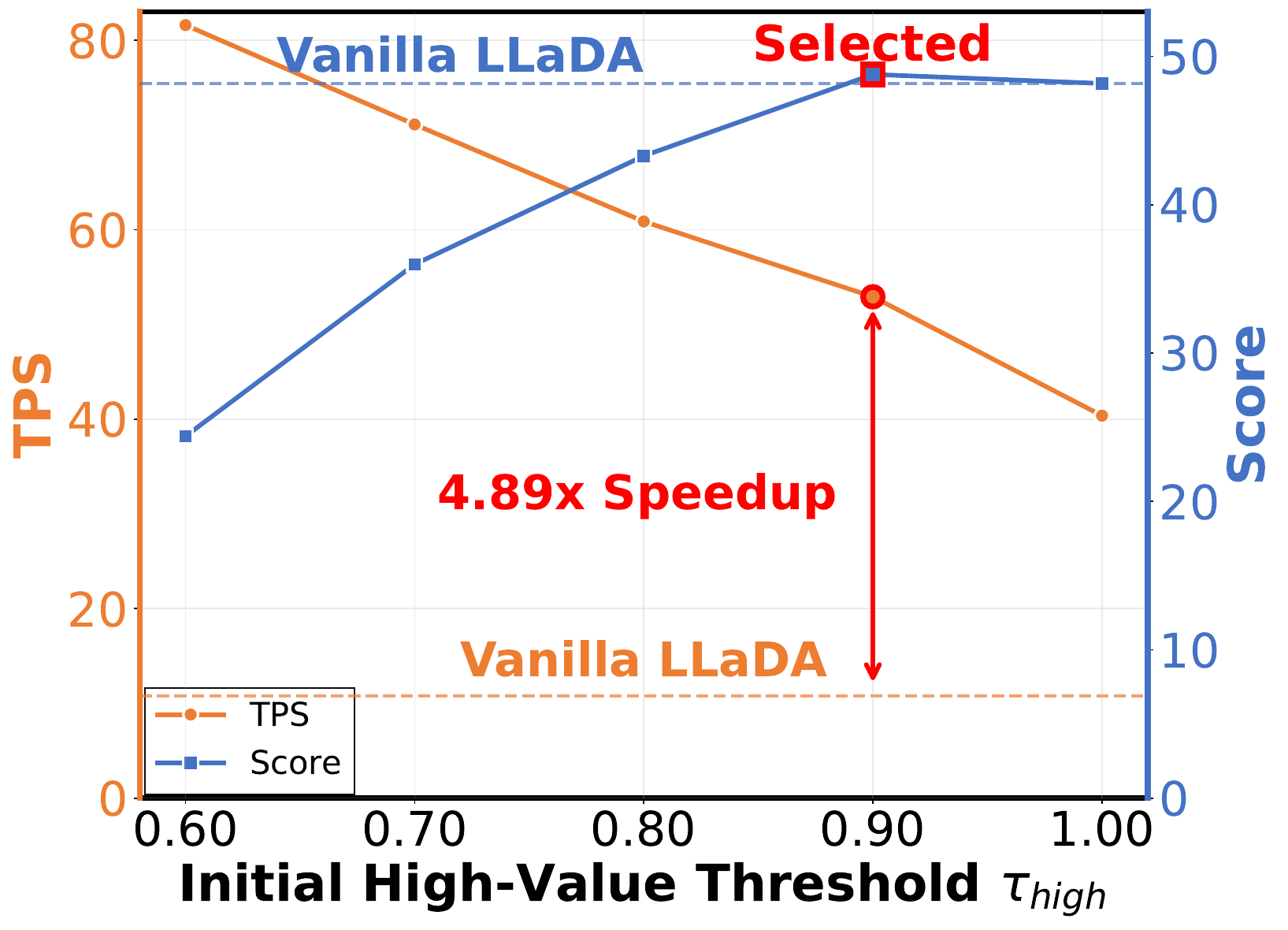}
  \caption{}
  \label{fig:masked-threshold}
\end{subfigure}
\begin{subfigure}[b]{0.244\textwidth}
  \centering
  \includegraphics[width=\textwidth]{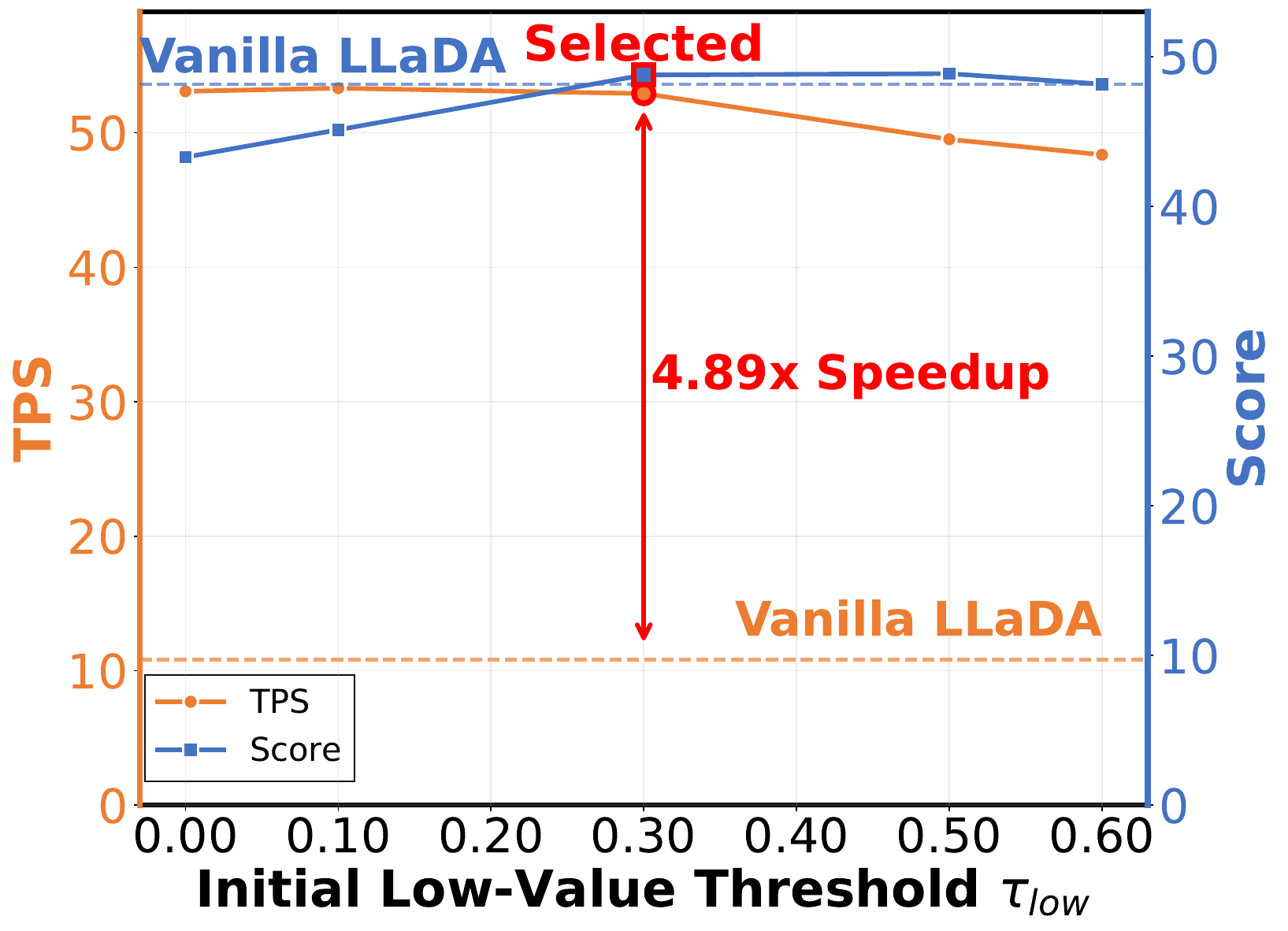}
  \caption{}
  \label{fig:decoded-threshold}
\end{subfigure}
\hfill
\begin{subfigure}[b]{0.244\textwidth}
  \centering
  \includegraphics[width=\textwidth]{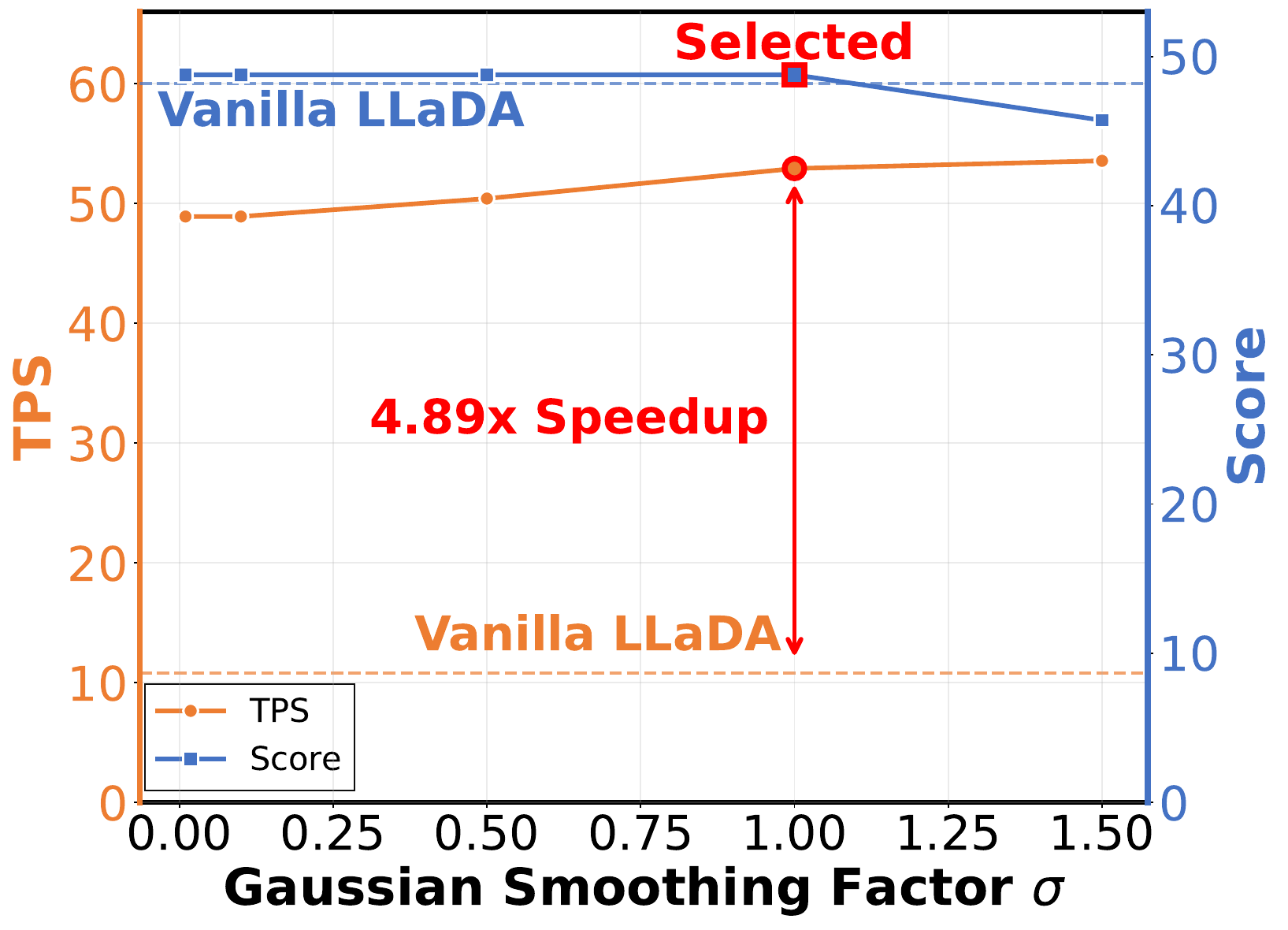}
  \caption{}
  \label{fig:sigma}
\end{subfigure}
\begin{subfigure}[b]{0.244\textwidth}
  \centering
  \includegraphics[width=\textwidth]{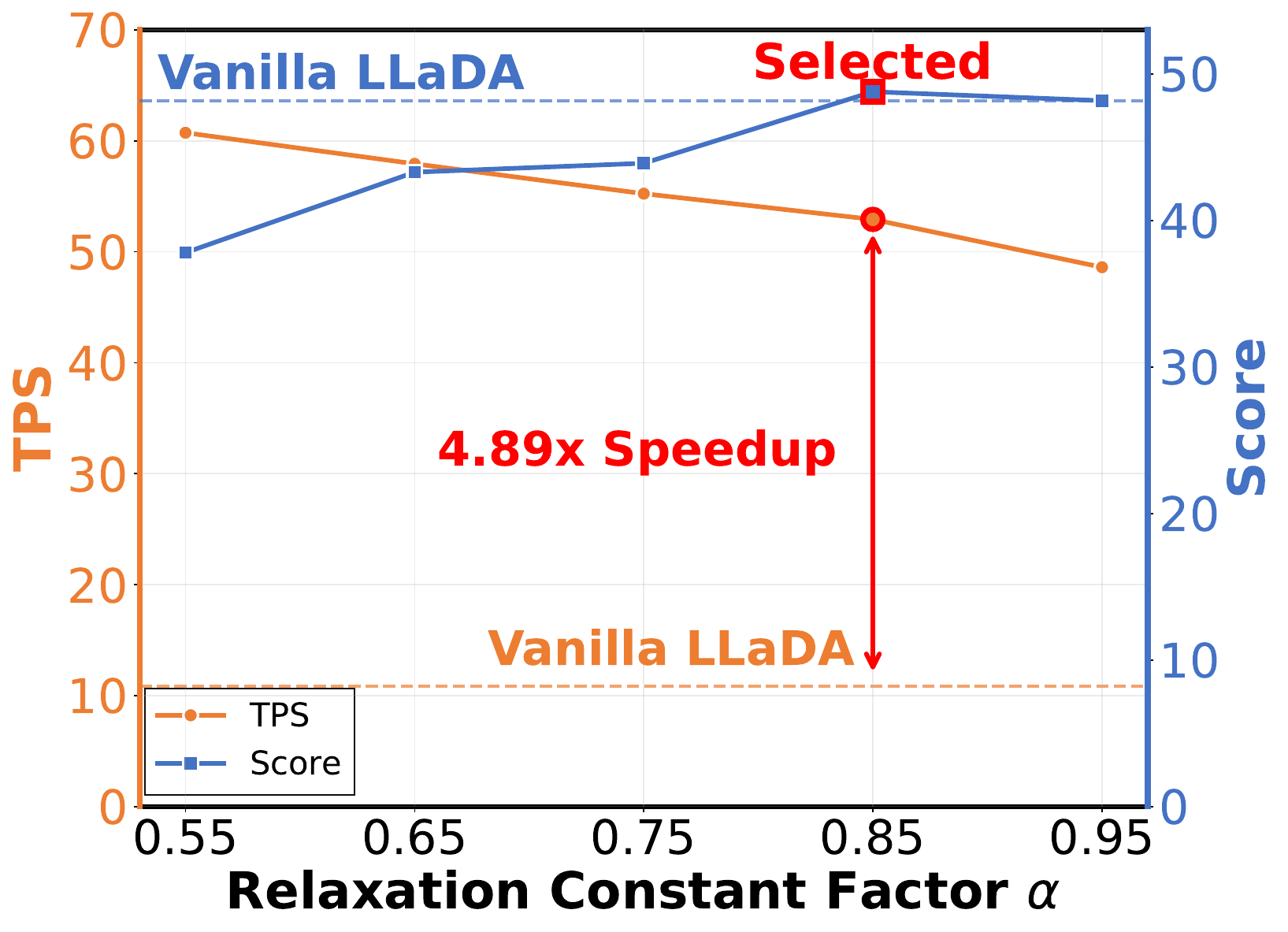}
  \caption{}
  \label{fig:relax}
\end{subfigure}
\caption{\textbf{Impact of hyperparameters in our \modelname{}}, including initial high-value thresholds $\tau_{\text{high}}$, initial low-value thresholds $\tau_{\text{low}}$, Gaussian smoothing factor $\sigma$, and the relaxation constant factor $\alpha$. The red markers denote our default setting, which achieves a $4.89\times$ speedup over vanilla LLaDA.} 
\vspace{-17pt}
\label{fig:base-threshold}
\end{figure}

\noindent\textbf{Ablation on Spatial-aware Decoding.} In spatial-aware decoding, we first introduce an initial high-value threshold $\tau_{\text{high}}$ and an initial low-value threshold $\tau_{\text{low}}$ to generate an initial threshold map, and then employ a Gaussian smoothing operation with the factor of $\sigma$ to generate the final threshold map. (i) We first show the impact of $\tau_{\text{high}}$ in Fig.~\ref{fig:masked-threshold}. When $\tau_{\text{high}}=0.9$, it presents best trade-off between generation score and TPS. When $\tau_{\text{high}}$ is larger than 0.9, it does not improve generation score while reducing speed. When $\tau_{\text{high}}$ is smaller than 0.9, although TPS shows some improvements, the generation score decreases significantly. (ii) We further validate the impact of  $\tau_{\text{low}}$ in Fig.~\ref{fig:decoded-threshold}. It has best trade-off when $\tau_{\text{low}}=0.3$. (iii) Finally, we show the impact of $\sigma$ in Fig~\ref{fig:sigma}. It achieves  best trade-off when $\sigma=1.0$. When $\sigma$ is too large, the influence of surrounding decoded tokens increases, leading to a decline in performance. When $\sigma$ is too small, it presents insufficient acceleration because the information from surrounding decoded tokens is not fully taken into account.

\noindent\textbf{Error Risk Analysis.}
We assess whether \modelname{} introduces additional decoding errors relative to the vanilla decoder. 
(1) \modelname{} maintains high semantic consistency with vanilla decoding. Specifically,  the cosine similarity between output token embeddings of our \modelname{} and vanilla decoding is  99.1\% and 98.9\% on HumanEval~\cite{humaneval} and GSM8K~\cite{cobbe2021gsm8k}, respectively. (2) We evaluate the aggregate error-budget certificate on decoding steps in Appendix~\ref{app:theory}. 
Our \modelname{} satisfies the certificate condition in more than 90\% of decoding steps, with step-level satisfaction rates of 90.8\% on GSM8K~\cite{cobbe2021gsm8k} and 91.7\% on HumanEval~\cite{humaneval}. These results suggest that our \modelname{} accelerates decoding while largely preserving the reference decoding behavior.
\section{Conclusion}
\label{sec:conclusion}

In this paper, we present an analysis on spatial and temporal stability in dLLM decoding, and introduce a spatio-temporal stability guided decoding approach for dLLMs, named \modelname{}. Our \modelname{} includes a spatial-aware decoding and a temporal-aware decoding, which aim to leverage the information of decoded tokens and ID consistency to adjust decoding thresholds for masked tokens. We perform the experiments on textual reasoning and multimodal understanding benchmarks, which significantly demonstrates the efficiency of our \modelname{}.

\label{app:limitations}

\textbf{Limitations.}
\modelname{} is a training-free inference acceleration method that improves efficiency by decoding more tokens at each denoising step. 
Therefore, its acceleration is more pronounced for tasks requiring relatively long generations, while the benefit can be limited for tasks with very short outputs, such as simple yes/no questions. 
For such tasks, reducing the per-step denoising cost may be more important than increasing the number of decoded tokens per step. 
\bibliographystyle{plainnat}
\bibliography{custom}

%%%%%%%%%%%%%%%%%%%%%%%%%%%%%%%%%%%%%%%%%%%%%%%%%%%%%%%%%%%%

\appendix
\clearpage
\appendix

In the appendix, we provide model experiment details in Section~\ref{sec:evaluation}, further theoretical analysis in Section~\ref{app:theory}, algorithm details in Section~\ref{sec:decoding}, extension to multi-modal dLLMs details in Section~\ref{sec:extension}, more ablation experiments in Section~\ref{sec:additional-ablation}, and more case studies in Section~\ref{sec:case}.

\section{Experiment Details}
\label{sec:evaluation}

\subsection{Benchmark Details}
We conduct experiments on textual reasoning and multimodal understanding benchmarks. In textual reasoning tasks, a detailed overview of benchmarks, as described below:
\begin{itemize}
    \item \textbf{HumanEval}~\cite{humaneval} is a hand-written evaluation set of 164 Python programming problems for docstring-conditioned function synthesis. The problems assess language comprehension, algorithms, and mathematics.
    \item \textbf{MBPP}~\cite{mbpp} is a Python code-generation benchmark of entry-level tasks with natural-language prompts and unit-test evaluation. 
    Following the standard split used by common evaluation suites, we evaluate on its problem test set, where each prediction is judged by passing the provided tests.
    \item \textbf{GPQA}~\cite{rein2024gpqa} is a graduate-level science QA benchmark with expert-written 4-choice multiple-choice questions spanning biology, physics, and chemistry.
    \item \textbf{GSM8K}~\cite{cobbe2021gsm8k} is a curated dataset of 8.5K human-written grade-school math word problems with step-by-step natural-language solutions, split into 7.5K training and 1K test examples. 
    \item \textbf{MATH}~\cite{hendrycksmath2021} is a problem evaluation subset of the MATH benchmark. It is formed by uniformly sampling held-out problems from the MATH test split, and its subject and difficulty distributions are designed to remain representative of the original test set. 
\end{itemize}
In multimodal understanding tasks, we use three math reasoning benchmarks as:
\begin{itemize}
    \item \textbf{MathVerse}~\cite{zhang2024mathverse} is a visual math benchmark with 2,612 diagram-based problems spanning plane geometry, solid geometry, and functions; each problem is rewritten into six multimodal variants, yielding about 15K test samples to better probe whether MLLMs truly use the diagrams.
    \item \textbf{MathVision}~\cite{wang2024measuring} is a multimodal math reasoning benchmark of 3,040 competition-style problems paired with visual contexts; it spans 16 mathematical disciplines and 5 difficulty levels to enable diverse, fine-grained evaluation of LMMs.
    \item \textbf{MathVista}~\cite{lu2024mathvista} is a visual math reasoning benchmark with 6,141 examples, aggregated from 28 existing multimodal math-related datasets plus three new sets (IQTest, FunctionQA, and PaperQA).
\end{itemize}

\subsection{Evaluation Toolkits}
For textual reasoning tasks, we conduct evaluations with the OpenCompass~\cite{opencompass} toolkit on two recent dLLMs, Dream-7B-Instruct~\cite{dream} and LLaDA-8B-Instruct~\cite{llada}. For multimodal understanding tasks, we use LMMS-Eval~\cite{lmms_eval2024} toolkit and report results on LaViDa-Reason~\cite{lavida}.

\subsection{Evaluation Setup}

\noindent\textbf{Textual Reasoning Task.} We follow the original evaluation pipeline in LLaDA~\cite{llada} using OpenCompass~\cite{opencompass} for both LLaDA~\cite{llada} and Dream~\cite{dream}. We show the generation length, denoising steps and block size in Table~\ref{tab:all_benchmarks}.

\noindent\textbf{Multimodal Understanding Task.}
We re-implemented the evaluation using LMMS-Eval~\cite{lmms_eval2024} for LaViDa-Reason~\cite{lavida}. We follow all the hyperparameters set in the evaluation script, including the schedule and step ratio. We show the generation length, denoising steps and block size in Table~\ref{tab:all_benchmarks}.
\begin{table}[t]
\caption{\textbf{Evaluation Setup Across All Benchmarks.} We report the detailed configuration, including generation length $L$, denoising steps $T$, and block size $B$, used in the main experiments.}
\centering
\renewcommand{\arraystretch}{1.0}
\setlength{\tabcolsep}{6pt}
\resizebox{\columnwidth}{!}{%
\begin{tabular}{lccc|lccc}
    \toprule
    {\bf Benchmark} & {\bf Length $L$} & {\bf Steps $T$} & {\bf Block Size $B$} 
    & {\bf Benchmark} & {\bf Length $L$} & {\bf Steps $T$} & {\bf Block Size $B$} \\
    \midrule
    MathVerse  & 512 & 512 & 512 & HumanEval & 512 & 512 & 32 \\
    MathVision & 512 & 512 & 512 & GPQA      & 128 & 128 & 64 \\
    MathVista  & 512 & 512 & 512 & GSM8K     & 256 & 256 & 32 \\
    MBPP       & 512 & 512 & 32  & MATH      & 512 & 512 & 64 \\
    \bottomrule
\end{tabular}%
}
\label{tab:all_benchmarks}
\end{table}

\section{Decoding Safety Analysis}
\label{app:theory}

In this section, we provide a detailed analysis of conservative single-step for token-adaptive decoding rule.
Similar to Fast-DLLM~\cite{fast-dllm}, we adopt the joint-PMF idealization to analyze parallel decoding in MDMs.
For the decoded positions selected by our \modelname{} at one denoising step, we show that under an aggregate error-budget condition, the token-adaptive parallel decoding result agrees with a greedy sequential decoding operating on the same selected positions.
This result should be understood as a sufficient analysis evaluated on the realized decoded set, rather than an additional constraint enforced during decoding.

\begin{wrapfigure}{r}{0.48\textwidth}
    \centering
    \vspace{-10pt}
    \includegraphics[width=\linewidth]{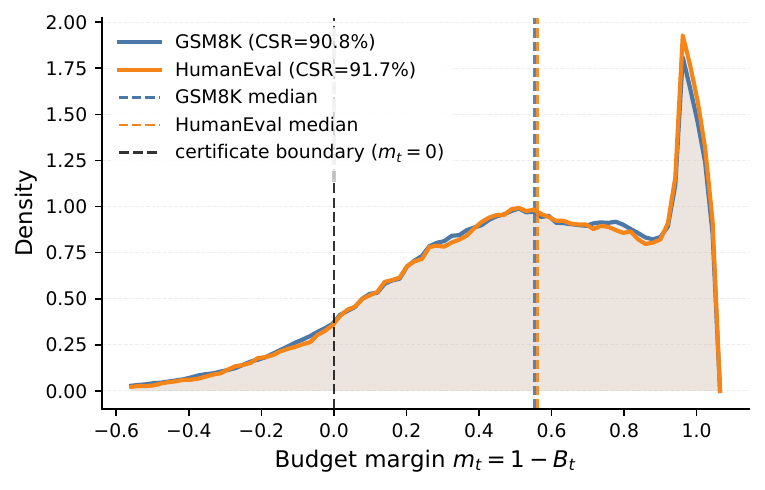}
    \caption{Budget-margin distribution over STDec decoding steps. Positive margins, computed as $m_t$, indicate satisfaction of the aggregate certificate condition in Eq.~\eqref{eq:app_aggregate_budget}. Most steps on both datasets fall in the positive region.}
    \label{fig:certificate_margin_distribution}
    \vspace{-18pt}
\end{wrapfigure}

\paragraph{Setup.}
At denoising step $t$, let $C_t=\{i_1,\ldots,i_n\}$ be the decoded set produced by the token-adaptive rule in Eq.~\ref{eq:commit_set_st}.
For each decoded position $i_j\in C_t$, the model predicts a marginal distribution $p_t(i_j)$ over the vocabulary.
Let $\hat{x}_t(i_j)=\arg\max_{v}p_t(i_j)[v]$, $c_t(i_j)=p_t(i_j)[\hat{x}_t(i_j)]$ be the top-1 prediction and its confidence.
We define the per-token error budget as $\varepsilon_j := 1-c_t(i_j)$.
The decoding rule $c_t(i)\ge \tau_t^{\mathrm{st}}(i)$ can equivalently be written as
$\varepsilon_t(i)\le 1-\tau_t^{\mathrm{st}}(i)$.
Following the joint-PMF idealization, assume that there exists a joint conditional distribution
$p(X_{i_1},\ldots,X_{i_n}\mid E_t)$ over the decoded positions whose marginals are the model-predicted distributions $p_t(i_j)$.
Here $E_t$ denotes the current decoded context and prompt.
We consider a greedy sequential decoder under the same joint PMF, and assume 
token-wise top-1 predictions as $x_t^\star=(\hat{x}_t(i_1),\ldots,\hat{x}_t(i_n))$ obtained by independently taking the top-1 prediction at each selected position.

\begin{theorem}[Conservative single-step analysis]
\label{thm:single_step_certificate}
Consider the decoded set $C_t=\{i_1,\ldots,i_n\}$ at step $t$, with the token-wise top-1 assignment
$x_t^\star=(\hat{x}_t(i_1),\ldots,\hat{x}_t(i_n))$ defined above.
If the aggregate error budget satisfies
\begin{equation}
\sum_{j=1}^{n}\varepsilon_j+\max_{1\le j\le n}\varepsilon_j<1,
\tag{B.1}
\label{eq:app_aggregate_budget}
\end{equation}
then $x_t^\star$ is the unique mode of the joint distribution over the decoded positions.
Therefore, under Eq.~\eqref{eq:app_aggregate_budget}, the token-adaptive parallel decoding result produced by \modelname{} agrees with greedy sequential decoding on the selected decoded set.
\end{theorem}

\begin{proof}
Let $S:=\sum_{j=1}^{n}\varepsilon_j,\varepsilon_{\max}:=\max_{1\le j\le n}\varepsilon_j$.
For each decoded position $i_j$, define the event
$A_j:=\{X_{i_j}=\hat{x}_t(i_j)\}$.
Since $\Pr(A_j^c\mid E_t)=1-c_t(i_j)=\varepsilon_j$, the union bound gives
\begin{equation}
\Pr(X=x_t^\star\mid E_t)
=
\Pr\left(\bigcap_{j=1}^{n}A_j \,\middle|\, E_t\right)
\ge
1-\sum_{j=1}^{n}\varepsilon_j.
\tag{B.2}
\label{eq:app_correct_lower}
\end{equation}

Now consider any assignment $z\neq x_t^\star$.
There exists at least one coordinate $k$ such that
$z_k\neq \hat{x}_t(i_k)$.
Therefore, $\{X=z\}\subseteq\{X_{i_k}=z_k\}$, and hence
\begin{equation}
\Pr(X=z\mid E_t)
\le
\Pr(X_{i_k}=z_k\mid E_t)
\le
\Pr(X_{i_k}\neq \hat{x}_t(i_k)\mid E_t)
=
\varepsilon_k
\le
\varepsilon_{\max}.
\tag{B.3}
\label{eq:app_wrong_upper}
\end{equation}

If Eq.~\eqref{eq:app_aggregate_budget} holds, then $1-S>\varepsilon_{\max}$.
Combining Eq.~\eqref{eq:app_correct_lower} and Eq.~\eqref{eq:app_wrong_upper}, we obtain
$p(x_t^\star\mid E_t)>p(z\mid E_t)$ for $z\neq x_t^\star$.
Thus, $x_t^\star$ is the unique mode of the joint distribution over the decoded tokens.

We prove by induction that the sequential decoder selects $\hat{x}_t(i_{\pi_k})$ at every step $k$.
For greedy sequential decoding under the same joint PMF, assume that before step $k$, the decoder has selected the corresponding top-1 entries of $x_t^\star$ for the previous positions. 
Let $P_{k-1}:=\bigcap_{\ell=1}^{k-1}\{X_{i_{\pi_\ell}}=\hat{x}_t(i_{\pi_\ell})\}$ denote this prefix event, with $P_0$ being the whole sample space.
Since $\{X=x_t^\star\}\subseteq P_{k-1}$, we have $\Pr(P_{k-1}\mid E_t)\ge 1-S>0$.
For the correct token, $\Pr(P_{k-1}\cap\{X_{i_{\pi_k}}=\hat{x}_t(i_{\pi_k})\}\mid E_t)\ge 1-S$.
For any competing token $v\neq \hat{x}_t(i_{\pi_k})$, 
$\Pr(P_{k-1}\cap\{X_{i_{\pi_k}}=v\}\mid E_t)\le \varepsilon_{\max}$.
Because Eq.~\eqref{eq:app_aggregate_budget} implies $1-S>\varepsilon_{\max}$, after dividing by the same positive denominator $\Pr(P_{k-1}\mid E_t)$, the conditional probability of $\hat{x}_t(i_{\pi_k})$ is strictly larger than that of any competing token.
Thus, greedy sequential decoding selects $\hat{x}_t(i_{\pi_k})$ at step $k$.
By induction, it returns $x_t^\star$.
Hence, the token-adaptive parallel decoding result produced by \modelname{} agrees with greedy sequential decoding on $C_t$.
\end{proof}

\paragraph{Empirical coverage.}
To measure how often realized \modelname{} decoding steps fall into this certified regime, we compute, for each decoded set,
\[
B_t=
\sum_{i\in C_t}(1-c_t(i))
+
\max_{i\in C_t}(1-c_t(i)),
\qquad
m_t=1-B_t.
\]
A positive margin $m_t>0$ indicates that the aggregate condition in Eq.~\eqref{eq:app_aggregate_budget} is satisfied.
In our experiments, \modelname{} reaches step-level certificate satisfaction rates of $90.8\%$ on GSM8K and $91.7\%$ on HumanEval, showing that most decoding steps fall within the certificate-covered regime.

\section{Algorithm Details of \modelname{}}
\label{sec:decoding}

Algorithm~\ref{alg:method} outlines the overall decoding pipeline. 
At each denoising step, our \modelname{} generates token-adaptive thresholds from spatial and temporal stability cues, enabling tokens with sufficiently stable predictions to be decoded earlier while preserving generation quality. To avoid rare deadlock steps where no token meets the threshold, we additionally force decoding of the most confident masked token (Line~17--19 in Algorithm~\ref{alg:method}), ensuring monotonic progress.
\begin{table}[t]
\caption{\textbf{Comparison on image generation.} We integrate our method into Lumina-DiMOO~\cite{lumina-dimoo} for text-to-image generation.}
\centering
\renewcommand{\arraystretch}{1.0}
\setlength{\tabcolsep}{4pt}
\begin{tabular}{l|ccc|ccc}
    \toprule
    \multirow{2}{*}{{\bf Method}} 
    & \multicolumn{3}{c|}{{\bf GenEval}} 
    & \multicolumn{3}{c}{{\bf DPGBench}} \\
    \cmidrule(lr){2-4} \cmidrule(lr){5-7}
    & {\bf TPS$\uparrow$} & {\bf Speed$\uparrow$} & {\bf Score$\uparrow$}
    & {\bf TPS$\uparrow$} & {\bf Speed$\uparrow$} & {\bf Score$\uparrow$} \\
    \midrule
    Lumina-DiMOO 
    & 50.47 & 1.00$\times$ & 0.85
    & 49.50 & 1.00$\times$ & 83.38 \\
    \cellcolor{gray!10}\textbf{+ \modelname{}}
    & \cellcolor{gray!10}\textbf{178.41}
    & \cellcolor{gray!10}\textbf{3.53}$\times$
    & \cellcolor{gray!10}0.87
    & \cellcolor{gray!10}\textbf{121.92}
    & \cellcolor{gray!10}\textbf{2.46}$\times$
    & \cellcolor{gray!10}85.28 \\
    \bottomrule
\end{tabular}%

\label{tab:lumina_ours_bench}
\end{table}

\begin{algorithm}[t]
  \caption{\modelname: Spatio-Temporal Stability Guided Decoding}
  \label{alg:method}
  \begin{algorithmic}[1]
  \REQUIRE model $p_\theta$, prompt $P$, generation length $L$, steps $T$
  \STATE Initialize masked sequence $X_T \gets \{\texttt{[MASK]}\}^L$, mask set $M_T \gets \{1,\dots,L\}$
  \STATE Initialize ID history $\hat{x}_{T+1}(i)\gets\texttt{NULL}$, temporal stability streak $K_{T+1}(i)\gets 0$ for all $i$
  \FOR{$t=T$ \TO $1$}
  \IF{$M_t=\emptyset$} \STATE $X_0 \gets X_t$ \STATE \textbf{break} \ENDIF
      \STATE Run $p_\theta$ on $(X_t,P)$ to obtain $\{p_t(i)\}_{i\in M_t}$
      \FOR{each $i\in M_t$}
          \STATE $\hat{x}_t(i)\gets\arg\max_{v}p_t(i)[v]$;\ \ $c_t(i)\gets p_t(i)\big[\hat{x}_t(i)\big]$
          \STATE $K_t(i)\gets K_{t+1}(i)+1$ if $\hat{x}_t(i)=\hat{x}_{t+1}(i)$, otherwise $K_t(i)\gets 0$
      \ENDFOR
      \STATE $\tau_t^{\text{init}}(i)\gets\tau_{\text{high}}$ for $i\in M_t$, and $\tau_t^{\text{init}}(i)\gets\tau_{\text{low}}$ for $i\notin M_t$
      \STATE $\tau_t^{s}(i)\gets f_G(\tau_t^{\mathrm{init}},\sigma)(i)$ for all $i\in M_t$
      \STATE $r_t(i)\gets \alpha$ if $K_t(i) > 1$,  $c_{t+1}(i)\ge \tau_{\text{low}}$, otherwise $r_t(i)\gets 1$
      \STATE $\tau_t^{\text{st}}(i)\gets r_t(i)\,\tau_t^{\text{s}}(i)$ for all $i\in M_t$
      \STATE $\mathcal{C}_t \gets \{\,i\in M_t \mid c_t(i)\ge \tau_t^{\text{st}}(i)\,\}$
      \IF{$|\mathcal{C}_t|=0$}
          \STATE $\mathcal{C}_t \gets \{\arg\max_{i\in M_t} c_t(i)\}$
      \ENDIF
      \STATE $M_{t-1}\gets M_t\setminus \mathcal{C}_t$, $X_{t-1}\gets X_t$
      \STATE Set $X_{t-1}(i)\gets \hat{x}_t(i)$ for $i\in\mathcal{C}_t$, and keep $X_{t-1}(i)\gets \texttt{[MASK]}$ for $i\in M_{t-1}$
  \ENDFOR
  \RETURN $X_0$
  \end{algorithmic}
  \end{algorithm}

\section{Extension to multi-modal dLLMs}
\label{sec:extension}

Our \modelname{} extends from 1D text sequences to 2D image lattices by leveraging the same spatio-temporal stability prior.
Concretely, at each denoising step, we build a pixel-wise adaptive threshold map by combining spatial cues from neighboring resolved pixels with temporal consistency of predicted pixel tokens across steps, and decode grid tokens accordingly.
We plug \modelname{} into the decoding procedure of Lumina-DiMOO~\cite{lumina-dimoo} and evaluate on standard text-to-image benchmarks (Table~\ref{tab:lumina_ours_bench}).
\modelname{} speeds up generation by $3.53\times$ on GenEval and $2.46\times$ on DPGBench, while keeping quality scores comparable, demonstrating its effectiveness for text-to-image generation.

\section{Additional Ablation}
\label{sec:additional-ablation}

\subsection{Ablation of Components in our \modelname{} on GSM8K}

We conduct additional ablation studies on GSM8K with LLaDA-8B-Instruct to further validate the robustness of our \modelname{}. 

\begin{wrapfigure}{r}{0.32\textwidth}
    \centering
    \vspace{-15pt}
    \includegraphics[width=\linewidth]{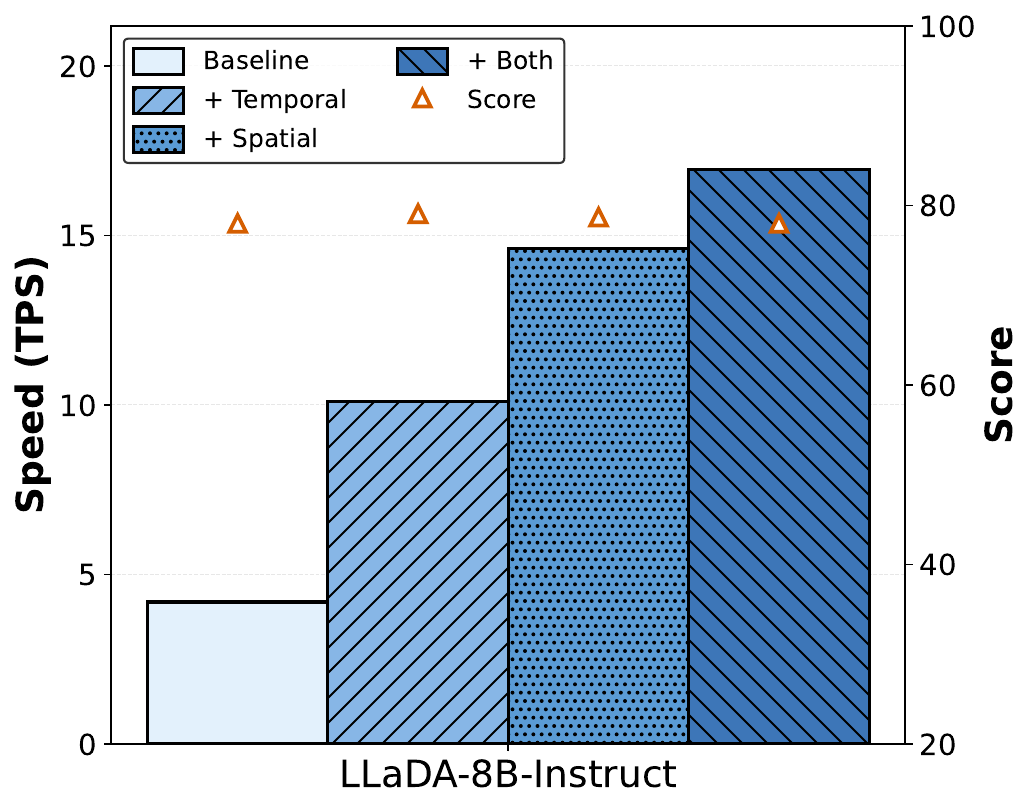}
    \caption{\textbf{Impact of integrating our \modelname{} modules on GSM8K.} We integrate our spatial-aware decoding and temporal-aware decoding into the baseline LLaDA.}
    \label{fig:ablation_module_gsm8k}
    \vspace{-25pt}
\end{wrapfigure}

\noindent\textbf{Module Ablation.}
Fig.~\ref{fig:ablation_module_gsm8k} shows that both components contribute to acceleration on GSM8K.
Temporal-aware decoding increases throughput from 4.19 to 10.1 TPS ($2.41\times$) with a small score gain (78.01$\rightarrow$79.08), while spatial-aware decoding reaches 14.61 TPS ($3.49\times$) with comparable score (78.7).
Combining both yields the best throughput (16.94 TPS, $4.04\times$) while preserving the baseline score (78.01), confirming their complementarity.

\begin{figure}[t]
\centering
\footnotesize
\begin{subfigure}[b]{0.245\textwidth}
  \centering
  \includegraphics[width=\textwidth]{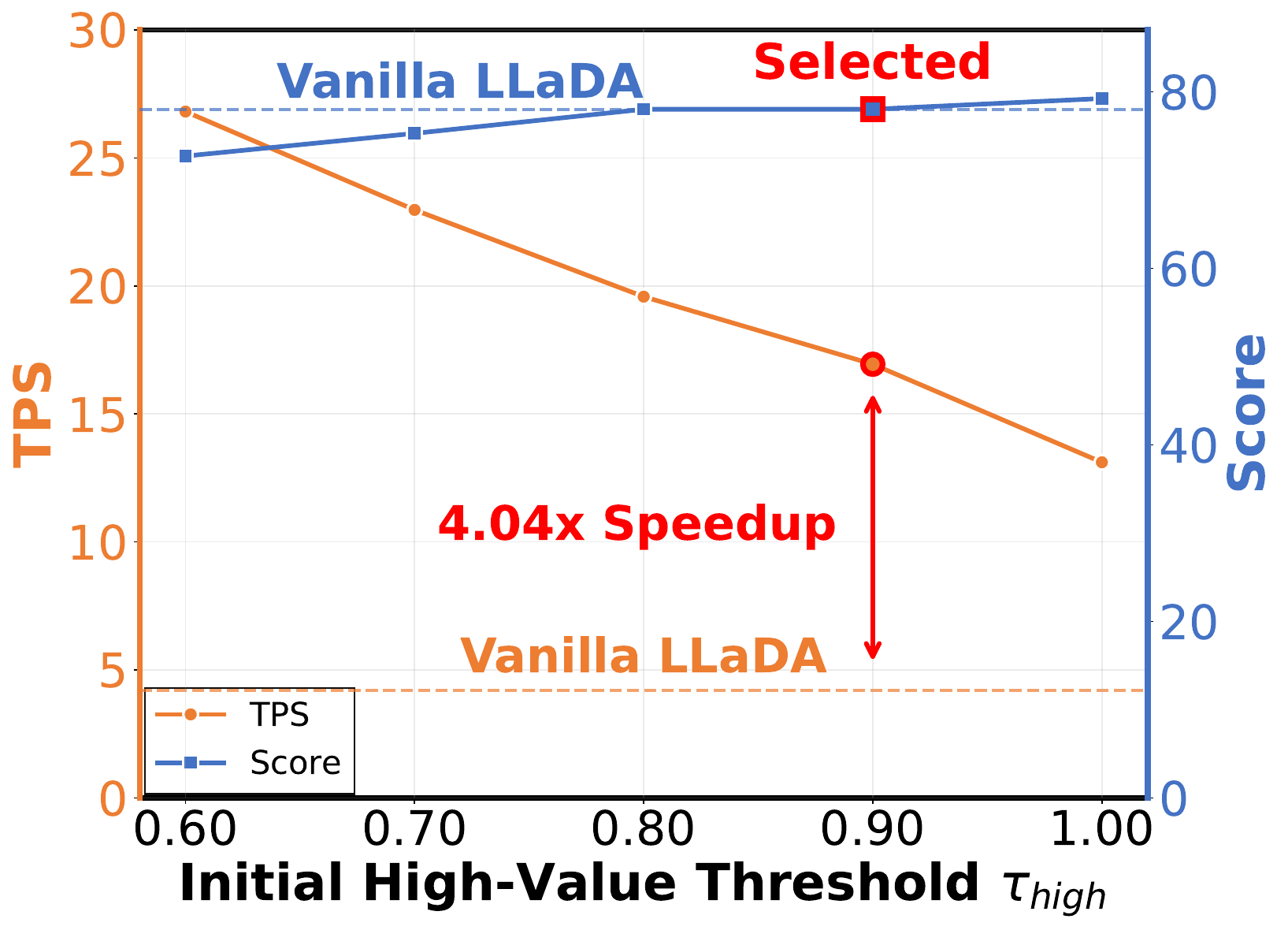}
  \caption{}
  \label{fig:masked-threshold_gsm8k}
\end{subfigure}
\begin{subfigure}[b]{0.245\textwidth}
  \centering
  \includegraphics[width=\textwidth]{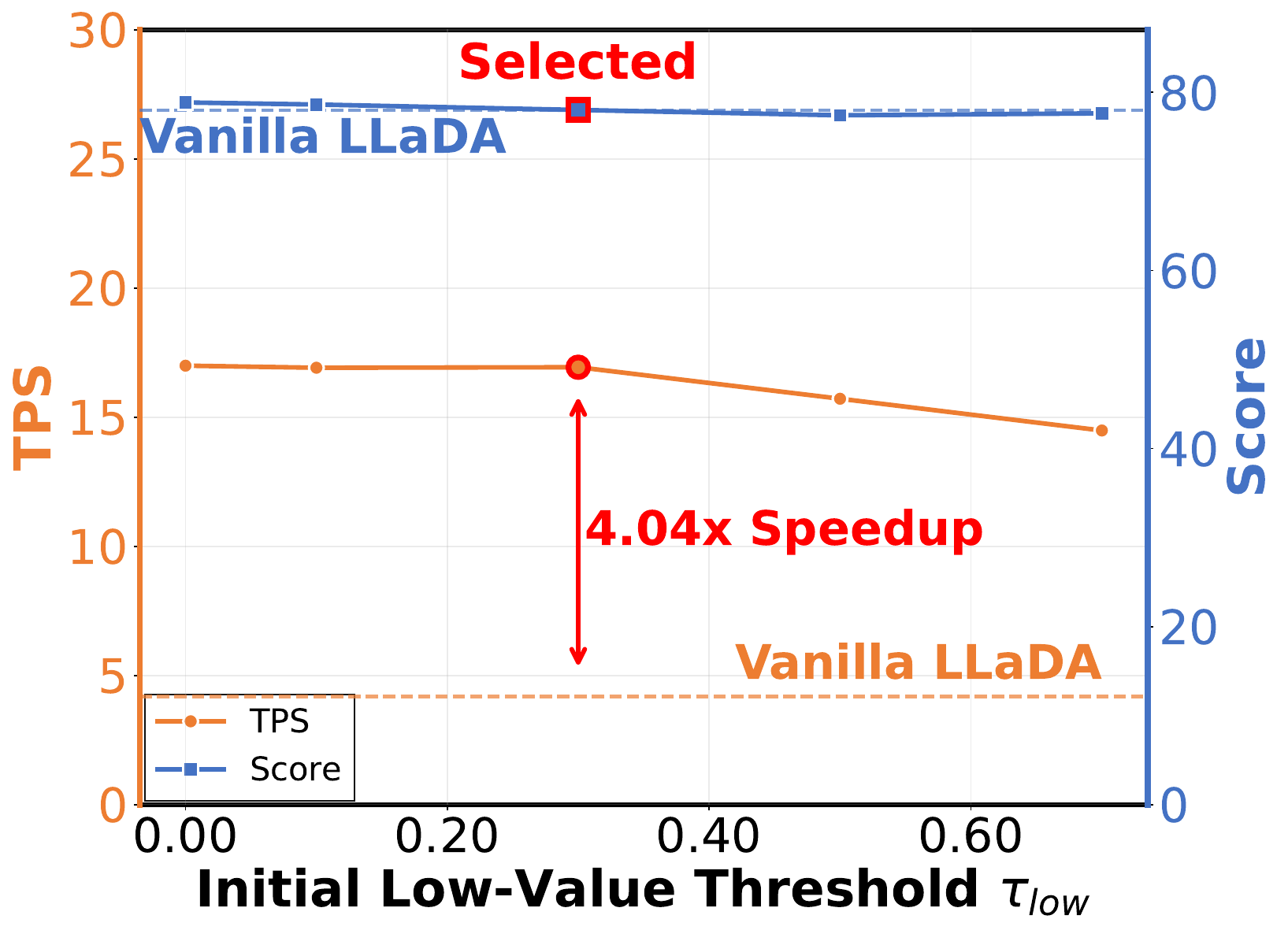}
  \caption{}
  \label{fig:decoded-threshold_gsm8k}
\end{subfigure}
\hfill
\begin{subfigure}[b]{0.245\textwidth}
  \centering
  \includegraphics[width=\textwidth]{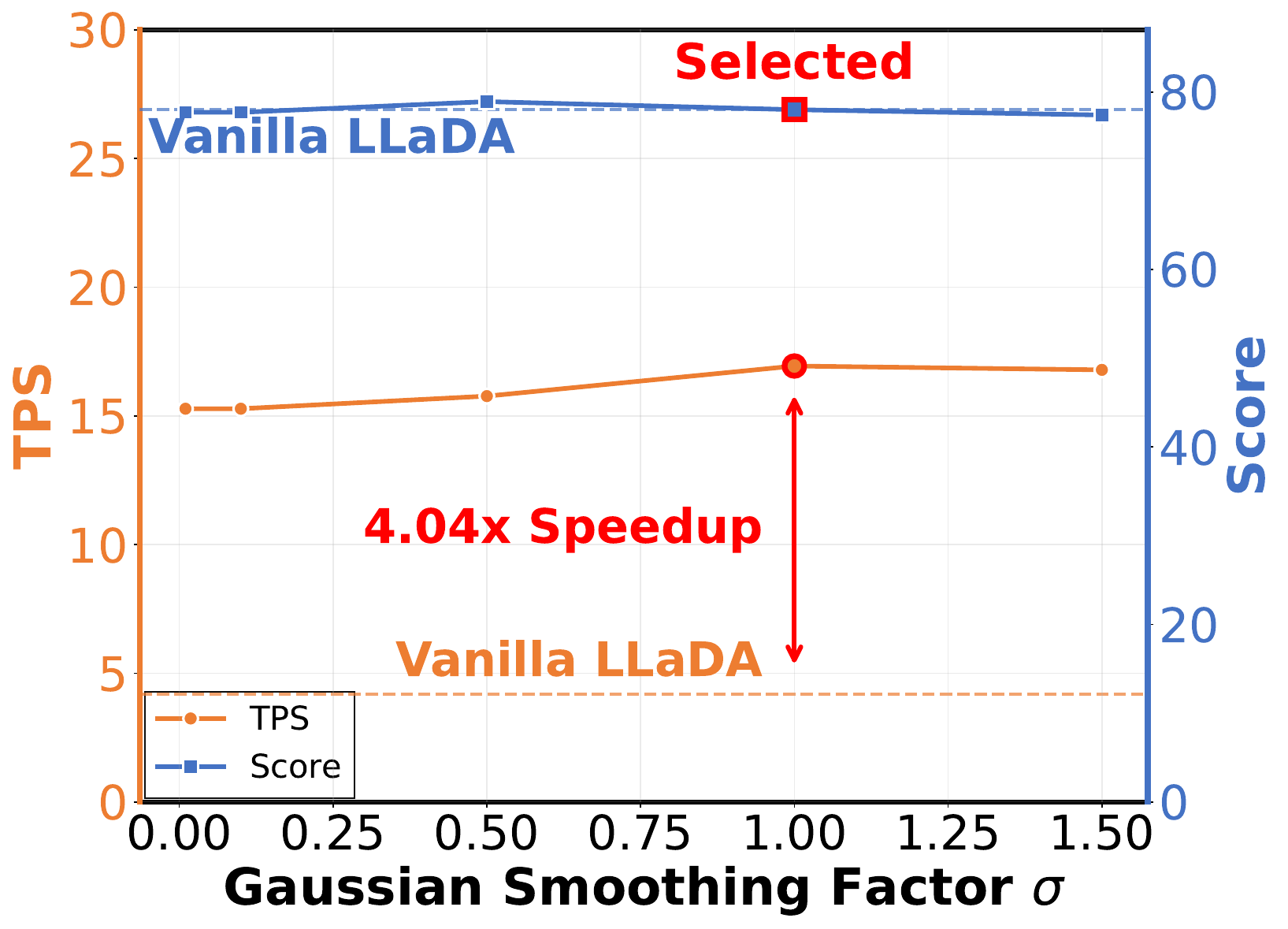}
  \caption{}
  \label{fig:sigma_gsm8k}
\end{subfigure}
\begin{subfigure}[b]{0.245\textwidth}
  \centering
  \includegraphics[width=\textwidth]{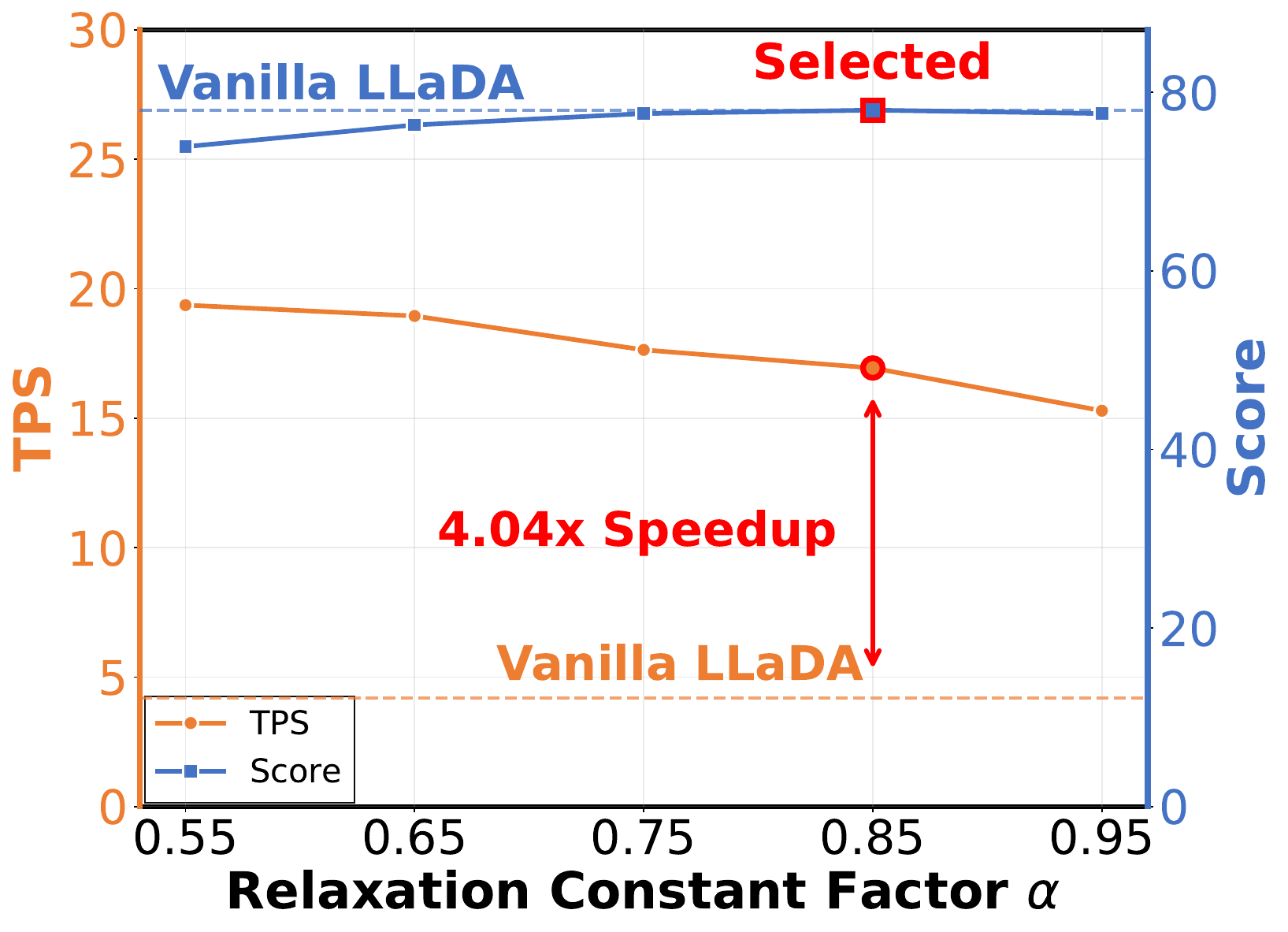}
  \caption{}
  \label{fig:relax_gsm8k}
\end{subfigure}
\caption{\textbf{Impact of different hyperparameters in our \modelname{} on GSM8K,}
including masked-token factor $\tau_{\text{high}}$, decoded-token factor $\tau_{\text{low}}$,
Gaussian smoothing factor $\sigma$, and the relaxation constant factor $\alpha$.
The red markers denote our default setting, which achieves a $4.04\times$ speedup over vanilla LLaDA.}
\label{fig:base-threshold-gsm8k}
\end{figure}

\noindent\textbf{Hyperparameter Robustness.}
Fig.~\ref{fig:base-threshold-gsm8k} reports the same hyper-parameter sweeps as in the main text.
The trends closely match our primary results: moderate $\tau_{\text{high}}$ and $\tau_{\text{low}}$ balance speed and quality, $\sigma{=}1.0$ remains a stable default, and $\alpha{=}0.85$ provides the strongest speed-quality trade-off.
Overall, \modelname{} is not sensitive to a narrow set of tuning choices, and the default setting transfers well to GSM8K.

\subsection{Smoothing Function for Spatial Decoding}
\label{app:kernel_ablation}

\begin{table}[t]
\caption{\textbf{Smoothing function ablation for spatial smoothing in \modelname{}.}
We compare different local smoothing kernels on GSM8K and HumanEval with LLaDA-8B-Instruct, reporting throughput (TPS), TPS speedup over the baseline, and task score.}
\centering
\footnotesize
\renewcommand{\arraystretch}{1.0}
\setlength{\tabcolsep}{7pt}
\resizebox{\columnwidth}{!}{%
\begin{tabular}{l|ccc|ccc}
    \toprule
    {\bf Smoothing Function} & \multicolumn{3}{c|}{{\bf GSM8K}} & \multicolumn{3}{c}{{\bf HumanEval}} \\
    \midrule
     & {\bf TPS$\uparrow$} & {\bf Speed$\uparrow$} & {\bf Score$\uparrow$}
     & {\bf TPS$\uparrow$} & {\bf Speed$\uparrow$} & {\bf Score$\uparrow$} \\
    \midrule
    Baseline   & 4.19  & 1.00$\times$ & 78.01 & 10.82 & 1.00$\times$ & 48.17 \\
    \midrule
    Mean       & 18.45 & 4.40$\times$ & 77.10 & 58.55 & 5.41$\times$ & 40.85 \\
    Triangular & 17.98 & 4.29$\times$ & 77.33 & 57.13 & 5.28$\times$ & 47.56 \\
    Gaussian   & 16.94 & 4.04$\times$ & 78.01 & 52.92 & 4.89$\times$ & 48.78 \\
    \bottomrule
\end{tabular}%
}
\label{tab:kernel_ablation_combined}
\end{table}
In fact, we can employ different smoothing functions used to smooth the initial threshold map in the spatial-aware module. We consider (i) a mean smoothing function that treats all neighbors within a radius $r$ equally. 
For each position $i$, we compute the spatial threshold $\tau^{\text{s}}(i)$ by averaging the initial thresholds in its local neighborhood as
\[
\tau^{\text{s}}(i) \;=\; \frac{1}{2r+1}\sum_{u=-r}^{r} \tau^{\text{init}}(i+u).
\tag{E.1}
% \label{eq:app_wrong_upper}
\]
Here, $r$ denotes the smoothing radius (window size $k=2r+1$), and $u$ indexes the discrete offset within the window. We also consider (ii) a triangular smoothing function that emphasizes closer neighbors with larger weights, which can be written as
\[
\begin{gathered}
\tau^{\text{s}}(i) \;=\; \sum_{u=-r}^{r} \alpha_u\, \tau^{\text{init}}(i+u), \\
\alpha_u \;=\; \frac{r+1-|u|}{\sum_{v=-r}^{r}(r+1-|v|)}.
\end{gathered}
\tag{E.2}
\]

Table~\ref{tab:kernel_ablation_combined}  reports the results of using different smoothing functions (Mean, Triangular and Gaussian). We observe that all three smoothing functions can accelerate inference speed, demonstrating that it is useful to leverage the neighboring information for efficient decoding. However, mean smoothing (distance-agnostic) leads to a notably larger score drop than triangular or Gaussian smoothing, suggesting that distance-aware weighting is important for preserving quality.
Namely, to enable more reliable decoding of tokens, smoothing operations need to consider distance information.

\section{Case Study}
\label{sec:case}

We conduct case studies on both textual reasoning and multimodal understanding tasks. In both cases, using our \modelname{} preserves the semantic content of the generated responses while substantially reducing decoding time.

\begin{figure*}[t]
\centering
\includegraphics[width=1.0\linewidth]{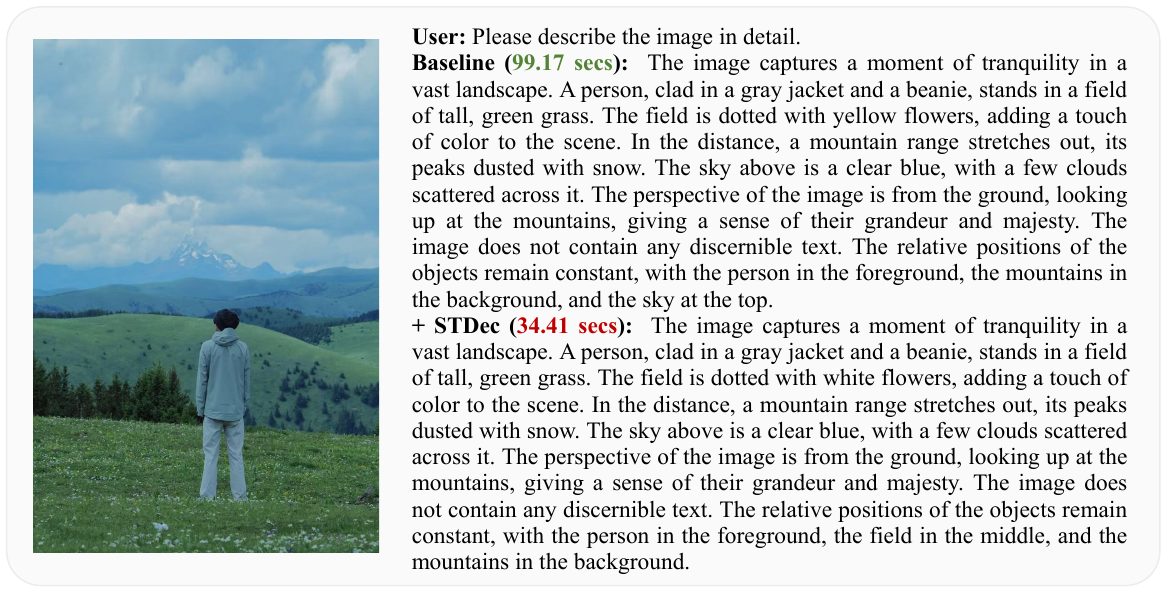}
\caption{\textbf{Case study on multimodal understanding with LaViDa-Reason~\cite{lavida}.} Given an input image, LaViDa-Reason generates a detailed scene description. Applying our \modelname{} preserves the semantic content and relative object relations in the response, while reducing decoding time from 99.17\,s to 34.41\,s ($2.88\times$ speedup).}

\label{fig:lavida-visualization}
\end{figure*}

\subsection{Textual Reasoning Task}
\begin{table*}[t]
\caption{\textbf{Case study on GSM8K~\cite{cobbe2021gsm8k} with Dream-7B-Instruct~\cite{dream}.} Our \modelname{} substantially reduces decoding latency while keeping the generated solution consistent with the baseline output.}
\label{tab:dream_vis}
\centering
\footnotesize
\renewcommand{\arraystretch}{1.0}
\setlength{\tabcolsep}{3pt}
\setlength{\parindent}{0pt}

{\renewcommand{\tabularxcolumn}[1]{>{\raggedright\arraybackslash}m{#1}}
\resizebox{\textwidth}{!}{%
\begin{tabularx}{\linewidth}{@{}>{\centering\arraybackslash}m{2.6cm}|X@{}}
\toprule
\textbf{Model} &
\textbf{Question 1:} Lily can run 12 kilometers per hour for 4 hours. After that, she runs 6 kilometers per hour. How many kilometers can she run in 8 hours?\\
\midrule

\makecell[c]{Dream-7B-Instruct\\[-1pt]\footnotesize 11.59 secs}
&%
Lily can run 12 kilometers per hour for 4 hours, so she can run 12 * 4 = 48 kilometers in those 4 hours.

After that, she runs 6 kilometers per hour for the next 4 hours, so she can run 6 * 4 = 24 kilometers in those 4 hours.

Thus, in 8 hours, she can run 48 + 24 = 72 kilometers.

So Lily can run \boxed{72} kilometers in 8 hours.\\
\midrule

\makecell[c]{+ \textbf{\modelname{}}\\[-1pt]\footnotesize \textbf{2.30 secs}~\textcolor{red}{(5.03$\times$)}}
&%
Lily can run 12 kilometers per hour for 4 hours, so she runs 12 * 4 = 48 kilometers in the first 4 hours.

After that, she runs 6 kilometers per hour for the next 4 hours, so she runs 6 * 4 = 24 kilometers in the next 4 hours.

Thus, in 8 hours, she can run 48 + 24 = 72 kilometers.

So Lily can run \boxed{72} kilometers in 8 hours.\\

% ===== Add 3 rows for Question 2 =====
\midrule
\multicolumn{2}{@{}l@{}}{\textbf{Question 2:} Joy can read 8 pages of a book in 20 minutes. How many hours will it take her to read 120 pages?}\\
\midrule

\makecell[c]{Dream-7B-Instruct\\[-1pt]\footnotesize 11.25 secs}
&%
To find out how many hours it will take Joy to read 120 pages, we first need to determine her reading speed in pages per minute.

Joy can read 8 pages in 20 minutes, so her reading speed is 8 pages / 20 minutes = 0.4 pages per minute.

Now, we need to find out how many minutes it will take her to read 120 pages. 

Since her reading speed is 0.4 pages per minute, it will take her 120 pages / 0.4 pages per minute = 300 minutes to read 120 pages.

Finally, we need to convert the time from minutes to hours. 

There are 60 minutes in an hour, so 300 minutes is equal to 300 minutes / 60 minutes per hour = 5 hours.

Thus, it will take Joy \boxed{5} hours to read 120 pages. \\

\midrule
\makecell[c]{+ \textbf{\modelname{}}\\[-1pt]\footnotesize \textbf{3.59 secs}~\textcolor{red}{(3.13$\times$)}}
&%
To find out how many hours it will take Joy to read 120 pages, we first need to determine her reading speed in pages per minute.

Joy can read 8 pages in 20 minutes, so her reading speed is 8 pages / 20 minutes = 0.4 pages per minute.

Now, we need to find out how many minutes it will take her to read 120 pages. We can do this by dividing the total number of pages by her reading speed:

120 pages / 0.4 pages per minute = 300 minutes.

Finally, we need to convert the time from minutes to hours. Since there are 60 minutes in an hour, we divide the number of minutes by 60: 

300 minutes / 60 minutes per hour = 5 hours.

So, it will take Joy \boxed{5} hours to read 120 pages.\\
\bottomrule
\end{tabularx}
}
}% end local tabularxcolumn redefinition

\end{table*}

\begin{table*}[t]
% \scriptsize
\footnotesize
\caption{\textbf{Case study on GSM8K~\cite{cobbe2021gsm8k} with LLaDA-8B-Instruct~\cite{llada}.} We apply our \modelname{} to speed up decoding while preserving the original reasoning and final answer.}
\label{tab:llada_vis}
\renewcommand{\arraystretch}{1.0}
\setlength{\tabcolsep}{3pt}
\setlength{\parindent}{0pt}

{\renewcommand{\tabularxcolumn}[1]{>{\raggedright\arraybackslash}m{#1}}

\begin{tabularx}{\linewidth}{@{}>{\centering\arraybackslash}m{2.6cm}|X@{}}
\toprule
\textbf{Model} &
\textbf{Question 1:} Lily can run 12 kilometers per hour for 4 hours. After that, she runs 6 kilometers per hour. How many kilometers can she run in 8 hours?\\
\midrule

\makecell[c]{LLaDA-8B-Instruct\\[-1pt]\footnotesize 12.04  secs}
&%
To determine how many kilometers Lily can run in 8 hours, we need to break down the problem into two parts: the distance she runs in the first 4 hours and the distance she runs in the next 4 hours.

First, let's calculate the distance she runs in the first 4 hours:
- She runs 12 kilometers per hour for 4 hours.
- Distance = Speed × Time
- Distance in the first 4 hours = 12 km/h × 4 h = 48 km

Next, let's calculate the distance she runs in the next 4 hours:
- She runs 6 kilometers per hour for 4 hours.
- Distance = Speed × Time
- Distance in the next 4 hours = 6 km/h × 4 h = 24 km

Now, we add the distances from both parts to find the total distance she can run in 8 hours:
- Total distance = Distance in the first 4 hours + Distance in the next 4 hours
- Total distance = 48 km + 24 km = 72 km

Therefore, Lily can run a total of 72 kilometers in 8 hours. \\
\midrule

\makecell[c]{+ \textbf{\modelname{}}\\[-1pt]\footnotesize \textbf{2.55 secs}~\textcolor{red}{(4.72$\times$)}}
&%
To determine how many kilometers Lily can run in 8 hours, we need to break down the problem into two parts: the distance she runs in the first 4 hours and the distance she runs in the next 4 hours.

First, let's calculate the distance she runs in the first 4 hours:
- Speed: 12 kilometers per hour
- Time: 4 hours
- Distance = Speed × Time
- Distance = 12 km/h × 4 h
- Distance = 48 kilometers

Next, let's calculate the distance she runs in the next 4 hours:
- Speed: 6 kilometers per hour
- Time: 4 hours
- Distance = Speed × Time
- Distance = 6 km/h × 4 h
- Distance = 24 kilometers

Now, we add the distances from both parts to find the total distance she can run in 8 hours:
- Total distance = Distance in the first 4 hours + Distance in the next 4 hours
- Total distance = 48 km + 24 km
- Total distance = 72 kilometers

Therefore, Lily can run **72 kilometers** in 8 hours.\\

% ===== Add 3 rows for Question 2 =====
\midrule
\multicolumn{2}{@{}l@{}}{\textbf{Question 2:} Joy can read 8 pages of a book in 20 minutes. How many hours will it take her to read 120 pages?}\\
\midrule

\makecell[c]{LLaDA-8B-Instruct\\[-1pt]\footnotesize 11.94 secs}
&%
To determine how many hours it will take Joy to read 120 pages, we need to follow these steps:

Step 1: Calculate Joy's reading rate in pages per minute.
Joy reads 8 pages in 20 minutes. Therefore, her reading rate is:
\[ \frac{8 \text{ pages}}{20 \text{ minutes}} = 0.4 \text{ pages per minute} \]

Step 2: Determine the total time required to read 120 pages.
Since Joy's reading rate is 0.4 pages per minute, the time required to read 120 pages is:
\[ \frac{120 \text{ pages}}{0.4 \text{ pages per minute}} = 300 \text{ minutes} \]

Step 3: Convert the total time from minutes to hours.
There are 60 minutes in an hour, so:
\[ \frac{300 \text{ minutes}}{60 \text{ minutes per hour}} = 5 \text{ hours} \]

Therefore, it will take Joy \boxed{5} hours to read 120 pages. \\

\midrule
\makecell[c]{+ \textbf{\modelname{}}\\[-1pt]\footnotesize \textbf{2.60 secs}~\textcolor{red}{(4.59$\times$)}}
&%
To determine how many hours it will take Joy to read 120 pages, we need to follow these steps:

Step 1: Calculate Joy's reading rate in pages per minute.
Joy reads 8 pages in 20 minutes. Therefore, her reading rate is:
\[ \text{Reading rate} = \frac{8 \text{ pages}}{20 \text{ minutes}} = 0.4 \text{ pages per minute} \]

Step 2: Determine the total time required to read 120 pages.
Since Joy reads at a rate of 0.4 pages per minute, the total time required to read 120 pages is:
\[ \text{Total time} = \frac{120 \text{ pages}}{0.4 \text{ pages per minute}} = 300 \text{ minutes} \]

Step 3: Convert the total time from minutes to hours.
\[ \text{Total time in hours} = \frac{300 \text{ minutes}}{60 \text{ minutes per hour}} = 5 \text{ hours} \]

Final answer: \boxed{5}\\
\bottomrule
\end{tabularx}
}

\end{table*}

Table~\ref{tab:dream_vis} and Table~\ref{tab:llada_vis} present examples on Dream~\cite{dream} and LLaDA~\cite{llada}. Compared with the vanilla model, our \modelname{} produces consistent final answers with similar reasoning content, but decodes noticeably faster by decoding locally stable tokens earlier.

\subsection{Multimodal Understanding Task}
Fig.~\ref{fig:lavida-visualization} shows a multimodal understanding example on LaViDa-Reason. Using our \modelname{} keeps the same key objects and their relative relations described by the baseline, while achieving a $2.88\times$ speedup (99.17\,s $\rightarrow$ 34.41\,s).

\end{document}